%% file: main.tex
\documentclass[twoside,11pt]{article}
\usepackage{jmlr2e} 
\usepackage{amsmath}
\usepackage{booktabs}
\usepackage{hyperref}
\hypersetup{hidelinks}
\usepackage{listings}
\usepackage{multirow}
\usepackage{array}
\usepackage{siunitx}
\usepackage{makecell}
\usepackage{float}

\ShortHeadings{Sequential Large Language Model-Based Hyper-parameter Optimization}{Mahammadli and Ertekin}
\firstpageno{1}

\begin{document}

\title{Sequential Large Language Model-Based Hyper-parameter Optimization}

\author{\name Kanan Mahammadli \email kanan.mahammadli@metu.edu.tr \\
       \addr Mathematics Department\\
       Middle East Technical University\\
       Ankara, Turkey
       \AND
       \name Seyda Ertekin \email
sertekin@metu.edu.tr \\
        \addr Computer Engineering Department\\
        \addr METU BILTIR CAD/CAM and Robotics Research Center\\
        Middle East Technical University\\
        Ankara, Turkey}
        
\editor{To Be Assigned}

\maketitle

\begin{abstract}%
This study introduces SLLMBO, an innovative framework leveraging large language models (LLMs) for hyperparameter optimization (HPO), incorporating dynamic search space adaptability, enhanced parameter space exploitation, and a novel LLM-tree-structured parzen estimator (LLM-TPE) sampler. By addressing limitations in recent fully LLM-based methods and traditional bayesian optimization (BO), SLLMBO achieves more robust optimization. This comprehensive benchmarking evaluates multiple LLMs—including GPT-3.5-Turbo, GPT-4o, Claude-Sonnet-3.5, and Gemini-1.5-Flash, extending prior work and establishing SLLMBO as the first framework to benchmark a diverse set of LLMs for HPO. By integrating LLMs' established strengths in parameter initialization with the exploitation abilities demonstrated in this study, alongside TPE’s exploration capabilities, the LLM-TPE sampler achieves a balanced exploration-exploitation trade-off, reduces API costs, and mitigates premature early stoppings for more effective parameter searches. Across 14 tabular tasks in classification and regression, the LLM-TPE sampler outperformed fully LLM-based methods and achieved superior results over BO methods in 9 tasks. Testing early stopping in budget-constrained scenarios demonstrated competitive performance, indicating that LLM-based methods generally benefit from extended iterations for optimal results. This work lays the foundation for future research exploring open-source LLMs, reproducibility of LLM results in HPO, and benchmarking SLLMBO on complex datasets, such as image classification, segmentation, and machine translation.
\end{abstract}


\input{sections/introduction}
\input{sections/literature_review}
\input{sections/methodology}
\input{sections/results_and_discussion}
\input{sections/conclusion}
\clearpage

\appendix
\input{appendices/appendix_A}

\input{appendices/appendix_B}

\vskip 0.2in
\bibliography{main}

\end{document}

%% file: sections/introduction.tex
\section{Introduction}
Machine learning models generally have two sets of parameters: model parameters and hyperparameters. Model parameters are learned during the training process, requiring no human interaction, while hyperparameters must be defined prior to training. Hyperparameter optimization (HPO) is critical for finding optimal solutions based on a given evaluation function \citep{tan2024hyperparameter1, wu2019hyperparameter2}.

\subsection{Manual and Bayesian Optimization}
The early stages of HPO required human experts with domain knowledge for the manual tuning process \citep{bergstra2011algorithms3, bergstra2012random4}, which is improved by sequential model-based global optimization (SMBO) methods \citep{bergstra2011algorithms3}. It has been shown that bayesian optimization is superior to other global optimization methods and is a state-of-the-art method for hyperparameter tuning \citep{tan2024hyperparameter1, wu2019hyperparameter2, hutter2019automated5}. However, it doesn’t fully automate the HPO process and has several limitations.\citep{wu2019hyperparameter2}.

\subsection{Limitations of Bayesian Optimization}
While BO offers significant advantages, it suffers from several drawbacks:
\begin{itemize}
    \item Human experts should define the parameters that need to be optimized and the possible set of values to choose from \citep{tan2024hyperparameter1, bergstra2011algorithms3}.
    \item Manually defined search space stays fixed throughout the optimization \citep{tan2024hyperparameter1, bergstra2011algorithms3}.
    \item Optimization can be too time-consuming through sequential process  \citep{tan2024hyperparameter1}.
    \item For every new task, bayesian optimization starts from scratch with random parameter initialization  \citep{feurer2014meta6, kim2017warm7, bai2023transfer8}.
\end{itemize}
Various ideas have been proposed to overcome the limitations of bayesian optimization, such as transfer-learned-based search spaces to reduce human effort \citep{shahriari2016unbounded9, nguyen2019filtering10}; designing algorithmic adaptive search spaces to solve fixed search space problems \citep{perrone2019search11, li2022transfer12}; multi-fidelity optimization \citep{hutter2019automated5, joy2020fast13} or early-stopping \citep{tan2024hyperparameter1} to speed up sequential learning; warm-starting to initialize parameters based on successful parameter values learned on similar datasets \citep{feurer2014meta6, kim2017warm7, bai2023transfer8, poloczek2016warm14}. However, each method suits a specific problem and does not lead to a unified framework. 

\subsection{Large Language Models for Hyperparameter Optimization}
Recent research has begun to explore using large language models to replace traditional random initialization or warm-starting and bayesian optimization methods in automated hyperparameter tuning  \citep{zhang2023llm15, liu2024agent16, liu2024bayesian17}. Preliminary studies indicate that LLMs can define the parameter search space through zero-shot learning without human expertise. Moreover, they may provide more effective initializations compared to random approaches. With few-shot learning, LLMs can perform optimization processes conditioned on the chat history of previous iterations, leveraging their internet-scale pretraining.

\subsection{Limitations of Current LLM-based Approaches and Lack of Research}
These studies have been conducted with only OpenAI’s models, can only run for 10 or 30 iterations based on the LLM prompting strategy due to the maximum input token limit and does not use adaptive search space methods. Hence, there are several important research questions regarding the usefulness of LLMs in reaching fully-automated HPO have yet to be answered:
\begin{itemize}
    \item How do different LLMs (GPT-3.5-turbo, GPT-4o, Claude-3.5-Sonnet, Gemini-1.5-Flash) perform in hyperparameter tuning compared to each other and bayesian optimization?
    \item Can summarization techniques extend the iteration limits imposed by input token constraints without compromising optimization quality?
    \item Can LLMs learn to update search space dynamically based on prior knowledge?
    \item Are large language models capable of autonomously managing the exploration-exploitation trade-off in optimization tasks without manually tuning control parameters?
    \item What strategies can be employed to automatically mitigate the risk of overexploitation in hyperparameter optimization processes that utilize large language models?
\end{itemize}

\subsection{Proposed Approach: Sequential Large Language Model-Based Optimization (SLLMBO)}
We propose SLLMBO, a novel method that leverages the strengths of LLMs and combines them with tree-structured parzen estimator sampling to address the identified research gaps. SLLMBO features an adaptive search space that evolves based on prior iterations and employs a hybrid LLM-TPE Sampler. It warm-starts with LLM and alternates between LLM-based sampling and TPE-based exploration to balance exploration and exploitation. Additionally, early stopping is integrated to terminate the optimization when no improvements are observed, thus reducing unnecessary computational costs.

\subsection{Contributions of This Study}
The key contributions of this study are as follows:
\begin{itemize}
    \item \textit{Comprehensive analysis of LLMs}: Benchmarking different LLMs against each other and bayesian optimization to evaluate their performance in hyperparameter optimization
    \item \textit{Adaptive search space}: A search space that evolves dynamically and automatically during optimization to focus on more promising regions.
    \item \textit{Hybrid LLM-TPE sampler}: A novel combination of LLM-based initialization and LLM-TPE sampling to enhance exploration-exploitation balance.
    \item \textit{Early stopping:} Integrated early stopping criteria to reduce computational costs and avoid overfitting during optimization.
\end{itemize}

%% file: sections/literature_review.tex
\section{Literature Review}
Hyperparameter optimization is critical to machine learning, as hyperparameters control the training process and significantly affect model performance. Unlike model parameters learned from the data, hyperparameters are predefined and must be set before training begins. Manual tuning is inefficient, error-prone, and often leads to non-reproducible results. This has led to automated HPO techniques, which aim to find optimal hyperparameters with minimal human intervention \citep{tan2024hyperparameter1, bergstra2011algorithms3, bergstra2012random4}. The increasing complexity of machine learning models has further motivated research into more sophisticated HPO methods \citep{tan2024hyperparameter1, bergstra2011algorithms3}.

\subsection{Classical HPO Methods}
The earliest methods for HPO, grid search, and random search, are widely used but have notable limitations. Grid search systematically explores all hyperparameter combinations in a predefined range. However, it becomes impractical due to the curse of dimensionality, where the number of evaluations grows exponentially with the number of hyperparameters \citep{tan2024hyperparameter1, wu2019hyperparameter2, bergstra2012random4, hutter2019automated5, hassanali2024software18}. Random search \citep{bergstra2012random4} improves efficiency by sampling hyperparameters randomly from the search space. It has been shown to outperform grid search in high-dimensional spaces \citep{bergstra2011algorithms3, bergstra2012random4, hutter2019automated5}. However, random search is unreliable for complex models, as it lacks structure and fails to leverage prior knowledge gained from previous evaluations \citep{wu2019hyperparameter2, bergstra2012random4, hutter2019automated5, hassanali2024software18}.

\subsection{Bayesian Optimization}
Bayesian optimization is a more advanced method that addresses the limitations of classical approaches by treating HPO as a black-box optimization problem \citep{snoek2012practical19}. BO aims to minimize an unknown objective function $f(x)$, where $f(x)$ represents the hyperparameter configuration. It builds a surrogate model of the objective function with probabilistic modeling based on previously evaluated hyperparameter configurations and uses an acquisition function to decide which configuration to evaluate next \citep{snoek2012practical19}. The bayesian theorem forms the core of BO, and the model is updated using new data as shown in Equation~\ref{eq:bayesian_update}:
\begin{equation}
P(\theta|D) \propto P(D|\theta)P(\theta)
\label{eq:bayesian_update}
\end{equation}
In Equation~\ref{eq:bayesian_update}, the posterior distribution $P(\theta|D)$ (updated belief about the hyperparameters given the data) is proportional to the product of the likelihood $P(D|\theta)$ (the probability of observing the data given the hyperparameters) and the prior $P(\theta)$ (initial belief about the hyperparameters). BO selects the next hyperparameter configuration by balancing exploration (sampling in less-known regions of the search space) and exploitation (sampling in regions known to perform well) using an acquisition function like expected improvement (EI), which chooses the next point based on the expected gain \citep{hutter2019automated5, snoek2012practical19}.

BO's most common surrogate model is the gaussian process (GP), which models the objective function as a distribution over functions. A GP provides a mean and variance for each point in the search space, offering uncertainty estimates exploited by the acquisition function to guide the search \citep{hutter2019automated5}. Despite its popularity, GPs have several limitations. Their computational complexity of fitting variances is $O(n^3)$, making them inefficient for large datasets. GPs also struggle with discrete and conditional hyperparameters, which limits their applicability in real-world problems \citep{tan2024hyperparameter1, hutter2019automated5}. 

Alternative surrogate models like the tree-structured parzen estimator (TPE) \citep{bergstra2011algorithms3} and sequential model-based algorithm configuration (SMAC) \citep{hutter2011sequential20} have been developed to overcome the limitations of GPs. TPE models the objective function by dividing the search space into two distributions with a predefined threshold $y^*$, as expressed in Equation~\ref{eq:tpe_distribution}:
\begin{equation}
P(x|y) = 
\begin{cases} 
l(x) & \text{if } y < y^* \\
g(x) & \text{if } y \geq y^*
\end{cases}
\label{eq:tpe_distribution}
\end{equation}
The optimization of the expected improvement for TPE is equivalent to maximizing the ratio shown in Equation~\ref{eq:tpe_ratio}:
\begin{equation}
\frac{g(x)}{l(x)}
\label{eq:tpe_ratio}
\end{equation}
SMAC extends BO to more complex hyperparameter spaces using Random Forests instead of GPs as the surrogate model \citep{hutter2011sequential20}. SMAC performs well in discrete and mixed search spaces and scales better than GPs. It also supports conditional hyperparameters, where specific hyperparameters only take effect if others are set to particular values \citep{hutter2019automated5}. Both TPE and SMAC methods have a computational complexity of $O(n \log n)$.

Despite using alternative surrogate models to reduce the computational cost of gaussian processes, several challenges remain in Bayesian Optimization. First, BO typically requires domain experts to design the search space, which must be carefully crafted to find the best parameters effectively \citep{tan2024hyperparameter1, bergstra2011algorithms3}. This reliance on predefined search spaces becomes problematic, as the search space remains static throughout the optimization, assuming it contains the global optimum \citep{perrone2019search11, li2022transfer12}. Poorly designed search spaces can miss optimal parameters, particularly in high-dimensional problems. Additionally, the optimization process usually begins with a random guess, which is often far from optimal, leading to inefficiency in the initial iterations and increased computational cost \citep{feurer2014meta6, kim2017warm7}. Finally, because BO is a black-box method, it may continue running even after finding the optimal solution, leading to unnecessary evaluations until the preset maximum number of iterations is reached \citep{tan2024hyperparameter1}. These limitations highlight the need for more adaptive and efficient techniques in BO.

Several methods have been proposed to address the limitations of bayesian optimization, though many remain task-specific and need to be fully generalizable. Dynamic search space techniques have been developed to overcome the problem of static search spaces. For example, some approaches propose gradually expanding the search space based on information gathered during optimization, allowing the process to begin in a smaller region and adapt over time \citep{shahriari2016unbounded9}. Other methods utilize filtering techniques that progressively expand the search space by filtering out less promising regions based on model performance \citep{nguyen2019filtering10}. While these strategies improve efficiency, they require user-defined boundaries and are only partially automated.

Warm-starting methods tackle the inefficiency of random initialization by leveraging successful hyperparameter configurations from previous tasks. These approaches use historical data to initialize the search, reducing the reliance on random guesses in the early stages \citep{kim2017warm7, poloczek2016warm14}. Transfer learning-based techniques refine this by designing compact search spaces based on prior optimizations, improving search efficiency in new tasks \citep{perrone2019search11, li2022transfer12}. However, these solutions depend on high-quality historical data and are limited when tasks need more apparent similarities.

Finally, early stopping techniques prevent unnecessary evaluations by terminating the optimization process when no significant improvements are observed over a predefined number of iterations. This method is broadly applicable and can effectively reduce computational costs, but it needs to address other challenges, such as search space design or parameter initialization \citep{tan2024hyperparameter1, hutter2019automated5, joy2020fast13}.

While these methods address specific challenges in Bayesian Optimization, they are generally not comprehensive or fully automated solutions, underscoring the need for a more unified approach.

\subsection{Large Language Models}
Large language models have become increasingly prominent, demonstrating exceptional capabilities across various domains, including education \citep{kasneci2023chatgpt21}, healthcare \citep{thirunavukarasu2023large22}, and academia \citep{meyer2023chatgpt23}. They have also shown great potential in tasks like code generation \citep{du2024evaluating24} and information retrieval \citep{zhu2023large25}. One of the key reasons for the effectiveness of LLMs is their ability to generalize to new tasks without fine-tuning, thanks to in-context learning \citep{wang2023learning26, wang2024latent27}. By providing LLMs with problem descriptions and relevant input-output examples (few-shot learning), these models can apply their pre-trained knowledge to novel tasks \citep{brown2020language28}. Moreover, LLMs can also perform zero-shot learning, solving tasks based solely on a problem description without needing examples \citep{kojima2022large29}.

LLMs have recently been explored for optimization tasks, where they are evaluated as potential replacements for traditional optimization methods. Studies have shown that LLMs can be effective in multi-objective evolutionary optimization through zero-shot learning \citep{liu2023large30} and combinatorial optimization using in-context learning \citep{liu2024large31}. Additionally, research has investigated using LLMs as surrogate models in black-box optimization for a molecular domain \citep{nguyen2024lico32}. These studies demonstrate that LLMs, with their ability to learn from context and apply knowledge across tasks, hold significant promise for optimization applications.

In hyperparameter optimization, the sequential nature of bayesian optimization aligns well with LLMs' in-context learning abilities. In BO, the initial step, akin to zero-shot learning, involves defining the search space and suggesting initial parameters based on the problem description. Subsequent iterations, where BO refines these parameters using feedback, resemble few-shot learning, where prior examples guide improved suggestions. This parallel suggests that LLMs, with their flexible learning approach, are well-positioned to handle the iterative nature of hyperparameter optimization.

While LLMs have demonstrated substantial potential across various optimization tasks, their application to hyperparameter optimization remains relatively unexplored. A few studies have begun to address this gap, offering promising insights and highlighting areas requiring further research to leverage LLMs' capabilities for HPO fully.

The first study \citep{zhang2023llm15} investigates GPT-3.5 and GPT-4 as replacements for traditional hyperparameter initialization and optimization techniques. The authors utilize the zero-shot learning capabilities of LLMs to suggest initial hyperparameter values based on problem descriptions and scikit-learn model docstrings, reducing the dependence on random initialization. Their evaluation across eight HPOBenchmark datasets reveals that LLMs perform comparably to random search and bayesian optimization when using few-shot learning for parameter updates within a constrained search budget of ten iterations. Additionally, they implement a chain-of-thought reasoning approach to extend the maximum iteration limit, where only the parameter-value pairs, rather than the entire chat history, are passed to the LLM. This method shows that LLMs using chain-of-thought reasoning perform similarly to the chat-based method, demonstrating the model's capacity to manage HPO tasks effectively, even with longer iterations.

AgentHPO \citep{liu2024agent16} introduces a two-agent framework—Creator and Executor—to automate hyperparameter tuning. The Creator agent, provided with a dataset, problem description, and a manually defined search space, suggests initial hyperparameter values. In contrast, the Executor agent conducts experiments and feeds backlog history to the Creator for subsequent optimization. The authors test AgentHPO over ten iterations across various tasks, including image classification, segmentation, tabular classification, regression, and machine translation. Their results indicate that AgentHPO surpasses random search and, in several cases, exceeds human benchmarks, demonstrating the effectiveness of LLMs in automatic tuning across diverse domains.

The third study, LLAMBO \citep{liu2024bayesian17}, proposes a novel approach that mimics the structure of Bayesian Optimization by replacing its key components with separate LLMs. LLAMBO includes a zero-shot warm starter, a surrogate model, and a candidate sampler, effectively reimagining the traditional BO pipeline. The authors evaluate LLAMBO against random search and Bayesian Optimization using different surrogate models: tree-structured parzen estimator, random forests, and gaussian processes, across 25 tasks from the Bayesmark benchmark. Their results show that LLAMBO outperforms these methods in all tasks based on average regret scores. Furthermore, LLAMBO introduces a surrogate model for exploration hyperparameters, allowing for a dynamic balance of the exploration-exploitation trade-off, which is optimized separately from the model hyperparameters.

Despite the advancements in recent studies, several significant limitations still exist in applying LLMs to hyperparameter optimization. Current research is primarily constrained by token limits, which restrict the number of iterations, and the impact of chat history summarization has yet to be fully explored. Additionally, these studies primarily rely on static search spaces, which, combined with the limited number of iterations allowed in optimization tasks, increase the risk of over-exploration without converging to optimal solutions. Dynamic search space adaptation, which could leverage prior knowledge for better exploitation and faster convergence, must be adequately investigated. Moreover, while LLAMBO \citep{liu2024bayesian17} introduces partial control over the exploration-exploitation trade-off, it remains limited and lacks full automation. Another significant gap is the exclusive focus on only OpenAI’s LLMs, leaving unexplored the comparative performance of models like Gemini-1.5-Flash or Claude-3.5-Sonnet in HPO tasks. Moreover, these studies have not evaluated the potential issue of overfitting hyperparameters on unseen test sets, raising concerns about the generalizability of the optimal hyperparameters found by LLMs. These limitations indicate the need for more comprehensive approaches to harness the potential of LLMs in HPO fully.

This paper addresses the existing research gaps by proposing a fully automated LLM-based framework for hyperparameter optimization. The strategy introduces dynamic search space adaptation, allowing the optimization process to evolve automatically based on prior knowledge, and employs iteration summarization techniques to overcome token limitations. A hybrid LLM-TPE sampler enhances the exploration-exploitation trade-off, combining LLM-based initialization with Bayesian Optimization. This approach includes early stopping criteria to reduce unnecessary computations and evaluates the generalizability of optimized hyperparameters on unseen test sets, addressing potential overfitting. Together, these innovations deliver a comprehensive, automated solution to HPO, overcoming the fundamental limitations of previous studies.

%% file: sections/methodology.tex
\section{Methodology}
\label{methodology}
\begin{figure}[ht]
\centering
\includegraphics[width=\textwidth]{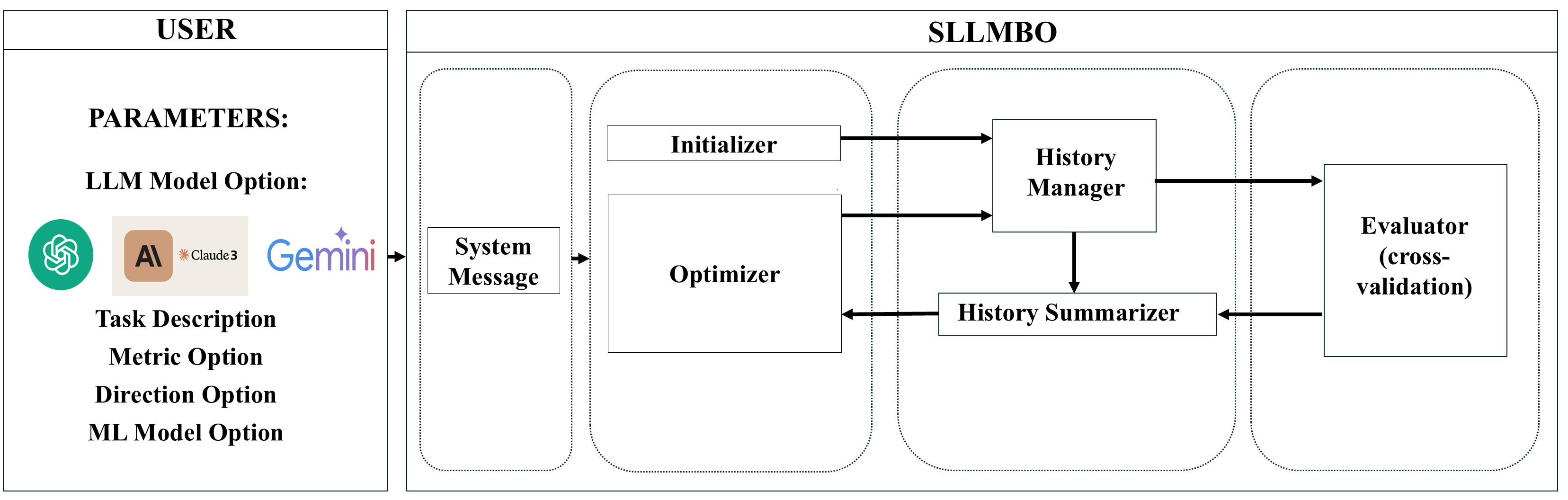}
\caption{The SLLMBO workflow. The framework consists of the Initializer, Optimizer, Evaluator, History Manager, and LLM-TPE Sampler components, working iteratively to perform efficient hyperparameter optimization.}
\label{fig:sllmbo_workflow}
\end{figure}

The SLLMBO framework, illustrated in Figure~\ref{fig:sllmbo_workflow}, is a fully automated hyperparameter optimization method that leverages the capabilities of large language models to guide the entire optimization process. SLLMBO interacts with the user-provided task description, followed by the automated generation of search spaces and parameter suggestions using zero-shot and few-shot learning. At the beginning of the process, the LLM initializes the search space and suggests initial hyperparameters. In subsequent iterations, the LLM dynamically updates the search space or keeps previously suggested ranges and suggests new parameters, informed by the history of prior iterations and the performance of the previously chosen values.

SLLMBO operates in iterative cycles between the following components: the Initializer, which is responsible for the first set of parameter ranges and values; the Optimizer, which refines parameter suggestions based on previous performance; the Evaluator, which calculates performance metrics through cross-validation; and the History Manager, which stores and summarizes previous interactions to prevent token limit overflows. Additionally, SLLMBO introduces a novel hybrid LLM-TPE Sampler, which dynamically balances exploration and exploitation by alternating between LLM-based and TPE-based sampling. Optional features such as reasoning-enabled optimization and early stopping further enhance the optimization process, making SLLMBO a robust and adaptive solution for complex HPO tasks.

\subsection{Initializer: Zero-Shot Parameter Initialization}
The Initializer is the starting point of the SLLMBO algorithm, where the LLM employs zero-shot learning to define the initial hyperparameter search space and suggest starting values. Based on the user-provided task description, machine learning model, evaluation metric, and optimization direction, the Initializer generates a structured JSON object containing the search space ranges and initial values. This allows the optimization process to begin with a meaningful exploration of the search space rather than random initialization, which is common in traditional methods. The system message and initialization prompt guide the LLM in performing this task.

\subsection{Optimizer: Few-Shot Learning for Dynamic Hyperparameter Updates}
Once the initialization is complete, the Optimizer component handles iterative updates of the search space and hyperparameter suggestions using few-shot learning. During each optimization cycle, the LLM is provided with feedback from the previous iterations, such as chat history and current best score, best parameter values, and the most recent search space in the optimization prompt. The LLM must decide whether to update the search space or continue with the current one, balancing exploration and exploitation.

In fully LLM-powered runs, the LLM suggests new parameters at every iteration. In contrast, in the LLM-TPE hybrid version, the LLM-TPE Sampler alternates between LLM-based suggestions and TPE-based sampling. This allows for a more diverse parameter space exploration, leveraging the LLM's generalization capabilities and TPE's statistical strengths.

\subsection{Evaluator: Cross-Validation for Metric Calculation}
The Evaluator is responsible for assessing the performance of each set of hyperparameters through cross-validation. In this study, 5-fold cross-validation was used, and the evaluation metric (e.g., F1-score or MAE) is passed back to the optimizer. This performance score guides the optimization process, with the LLM adjusting its future parameter suggestions based on how well the current set performs.

\subsection{History Manager: Memory Handling and Summarization}
The History Manager ensures that the optimization process maintains continuity across iterations by storing all parameter suggestions, evaluation scores, and decision points. The History module stores this information fully, while the Summarizer condenses the history when input token limits are reached. In the experiments, two summarization methods were used:
\begin{itemize}
    \item \textbf{Intelligent Summary}: An LLM-based method that generates concise summaries when prompted, after a certain number of iterations, or when the input token limit is exceeded.
    \item \textbf{LangChain Memory Buffer}: Used in LangChain-based experiments, this automatically manages memory without manual summarization, allowing for a seamless continuation of the optimization process.
\end{itemize}
Intelligent Summary is only used in fully LLM-empowered optimization with only GPT-3.5-turbo, and more details are given in the results and discussion section.

\subsection{LLM-TPE Sampler: Hybrid Sampling Strategy}
The LLM-TPE Sampler is a hybrid approach alternating between LLM-guided parameter suggestions and TPE-based sampling to balance exploration and exploitation. This sampler suggests the next hyperparameter value based on a custom function defined in Equation~\ref{eq:llm_tpe_sampler}:
\begin{equation}
S(X) = 
\begin{cases} 
\text{LLM}(X_\text{chat\_history}) & \text{if } p < 0.5 \\
\text{TPE}(X_\text{prev\_iters,search\_space}) & \text{otherwise}
\end{cases}
\label{eq:llm_tpe_sampler}
\end{equation}
In Equation~\ref{eq:llm_tpe_sampler}, $p \sim U(0,1)$, which means that half the iterations are guided by the LLM, using previous chat history and performance results, while the other half rely on TPE to explore regions of the search space that the LLM may not fully exploit. This dynamic approach ensures that SLLMBO avoids getting stuck in local optima and can effectively explore uncertain areas of the search space. The LLM-TPE sampler is designed as an alternative sampler for the optuna library.

\subsection{Reasoning-Enabled Optimization}
SLLMBO offers an optional Reasoning Engine to enhance transparency and interpretability. This feature requires the LLM to explain each decision made during the optimization process: whether to update or retain the current search space and why specific parameters were chosen. These explanations are stored alongside the optimization history, allowing for post-hoc analysis of the optimization decisions.

\subsection{Early Stopping Mechanism}
SLLMBO also incorporates an early stopping mechanism to prevent unnecessary iterations and reduce computational costs. The optimization process halts if the evaluation metric does not improve for a predefined number of iterations (five in our experiments). This mechanism is applied to the fully LLM-powered runs and the hybrid LLM-TPE method, ensuring efficiency and preventing overfitting in cases where further iterations are unlikely to yield better results.

Appendix~\ref{appendix_A} contains system messages and exact prompts used for initialization, optimization, summarization, and reasoning.

\subsection{Experimental Setup}
\label{methodology:experimental_setup}
The experimental evaluation of SLLMBO was conducted on six datasets: three classification and two regression tasks from The UCI Machine Learning Repository \citep{kelly2023uci33}, one regression task from OpenML \citep{vanschoren2014openml34}, and the M5 Forecasting Accuracy dataset \citep{howard2020m535}, a challenging multiple time series forecasting task. The classification datasets include gas\_drift \citep{vergara2012gas36}, cover\_type \citep{blackard1998covertype37}, and adult\_census \citep{becker1996adult38}, while the regression datasets include bike\_sharing \citep{fanaee2013bike39}, concrete strength \citep{vanschoren2014openml34}, and energy \citep{candanedo2017appliances40}.
For each dataset, approximately 20\% of the data is kept for testing the performance of the best parameters found by some HPO method on the rest of the data, helping to explore overfitting.
The F1-score was used for classification tasks to assess the performance of each iteration during the HPO process, while the mean absolute error (MAE) was employed for regression tasks. For parameter suggestion in subsequent iterations, the average cross-validation score across the folds was used as the performance metric. Detailed descriptions of each dataset, including feature and target column selection, preprocessing, feature engineering, and the specific cross-validation techniques applied, can be found in Appendix~\ref{appendix_B}.

The SLLMBO was compared against state-of-the-art Bayesian Optimization methods using two popular libraries: Optuna \citep{akiba2019optuna41} and Hyperopt \citep{bergstra2015hyperopt42}, both leveraging the Tree-structured Parzen Estimator (TPE) as their sampling strategy. The choice of Optuna and Hyperopt is grounded in recent studies highlighting their effectiveness and widespread use in tabular tasks \citep{hassanali2024software18, lai2023tree43, lv2023gdbt44, arden2022comparison45, putatunda2019bayesian46, wang2020lightgbm47, omkari2022cardiovascular48}. LightGBM and XGBoost were used as machine learning models, known for their superior performance on tabular datasets \citep{lai2023tree43, lv2023gdbt44, arden2022comparison45, wang2020lightgbm47, omkari2022cardiovascular48}.

The following configurations were evaluated for each dataset and model (LightGBM and XGBoost). The names provided below correspond to the unified naming conventions used in the tables and figures to ensure clarity:

\begin{itemize}
    \item \textbf{llm\_sampler+intelligent summary+gpt-3.5-turbo+patience\_15}: This first strategy uses a fully LLM-powered process with GPT-3.5-Turbo for initialization, optimization, and intelligent summarization to handle token limits. 
    \label{methodology:llm_strategy1}
    
    \item \textbf{llm\_sampler+langchain+\{\textit{optimizer}\}+patience\_15}: Second strategy replaces intelligent summarization with LangChain's memory buffer and \textit{optimizer} is one of the possible LLMs:
    \begin{itemize}
        \item GPT-4o
        \item GPT-3.5-Turbo
        \item Claude-3.5-Sonnet
        \item Gemini-1.5-Flash
    \end{itemize}
    \label{methodology:llm_strategy2}

    \item \textbf{llm-tpe sampler+\{\textit{optimizer}\}+\{\textit{initializer}\}+patience\_\{\textit{int}\}}: The last strategy uses a novel LLM-TPE Sampler instead of fully LLM-based optimization. \textit{optimizer} is one of GPT-4o and Gemini-1.5-Flash, and \textit{initializer} is either Random\_Init representing random warmstarting, or LLM\_Init where LLM suggests the first hyperparameters. The \textit{int} represents the integer early stopping for patience, which is either 15 as in previous strategies or 5 for testing the LLM-TPE Sampler with strict regularization.
    \label{methodology:llm_strategy3}
\end{itemize}

\subsubsection*{Additional Experimentation Details}
\begin{itemize}
    \item \textit{Number of iterations}: Each experiment was run for 50 iterations.
    \item \textit{Random seed}: A fixed random seed of 42 was used across all experiments to ensure reproducibility.
    \item \textit{Early Stopping Criteria}: In fully LLM-powered setups (both with Intelligent Summary and LangChain memory handler), an early stopping mechanism with a patience threshold of 15 iterations was employed to prevent unnecessary iterations and minimize API costs. In the LLM-TPE hybrid setups, a stricter early stopping threshold of 5 iterations was applied to assess performance under a limited budget and harsher regularization conditions. For configurations where early stopping was not explicitly used, the optimization ran for 50 iterations.
    \item \textit{Intelligent Summary}: A summarization step was triggered every ten iterations for the experiments, which utilizes intelligent summarization with GPT models to manage token limits and retain critical historical information for the next iterations.
\end{itemize}

This setup was designed to rigorously compare the performance of SLLMBO against traditional BO methods while evaluating the impact of different LLM models, memory handling mechanisms, hybrid LLM-TPE sampling, and early stopping of the optimization process.

%% file: sections/results_and_discussion.tex
\section{Results and Discussion}
The experiments were carried out sequentially to explore different strategies for hyperparameter optimization using LLM-powered methods, LLM-TPE hybrid samplers, and traditional Optuna/Hyperopt approaches. Each successive experiment was designed based on the findings of the previous one, which led to further refinement and informed decisions on the next phase of experimentation. This section details each process step's key findings, reasoning, and outcomes, followed by a discussion of the results and answers to the research questions posed. Comparing SLLMBO results with different strategies with each other and with Optuna and Hyperopt libraries using TPE as a surrogate model, each method is experimented on seven datasets, with both LightGBM and XGBoost models, making a total of 14 tasks.

\subsection{Initial Exploration: SLLMBO with Fully LLM-powered Methods and Intelligent Summary}
\label{subsection_4_1}

The initial experiments utilized the GPT-3.5-Turbo model for parameter initialization, optimization, and iterative summarization of the optimization history. The goal was to assess the effectiveness of the initial LLM strategy in HPO compared to established methods such as Optuna and Hyperopt. Tables~\ref{tab:table1} and~\ref{tab:table2} summarize the initialization, optimization, and runtime performance for classification and regression tasks, respectively.
\input{tables/table1.tex} 
\input{tables/table2.tex} 

Regarding parameter initialization, the LLM-powered approach consistently outperformed traditional random initialization in 9 out of 14 tasks (see Tables~\ref{tab:table1} and~\ref{tab:table2}). Notably, in 6 out of 8 regression tasks, the initial parameters suggested by the LLM yielded scores closer to the best score than those indicated by Optuna and Hyperopt, as summarized in Table~\ref{tab:table2}. This demonstrates the LLM's superior capability for parameter initialization, especially in regression tasks.

When evaluating the cross-validation (CV) and test scores of the best parameters found through optimization, the LLM-based HPO method also performed competitively. Across multiple tasks, it provided better or comparable results to both Optuna and Hyperopt. Specifically, the LLM surpassed Hyperopt in 5 tasks while achieving competitive results with Optuna in several tasks and two tasks better than Optuna and Hyperopt. In the energy dataset with XGBoost, Table~\ref{tab:table2} shows that the LLM-based approach produced a 17.5\% lower MAE on the test set than Optuna despite Optuna finding a slightly better cross-validation score. In contrast, in the energy dataset with LightGBM, the LLM-based method resulted in a 25\% higher test error (MAE) than Optuna. All optimization methods exhibited overfitting in the bike\_sharing and energy datasets, spanning four tasks. However, the LLM-based approach for the energy dataset with XGBoost showed noticeably softer overfitting than the other methods. These results suggest that while the LLM-powered HPO can perform strongly in regression and classification tasks, its effectiveness depends on the model and dataset characteristics.

\begin{figure}[!ht]
    \centering
\includegraphics[width=\textwidth]{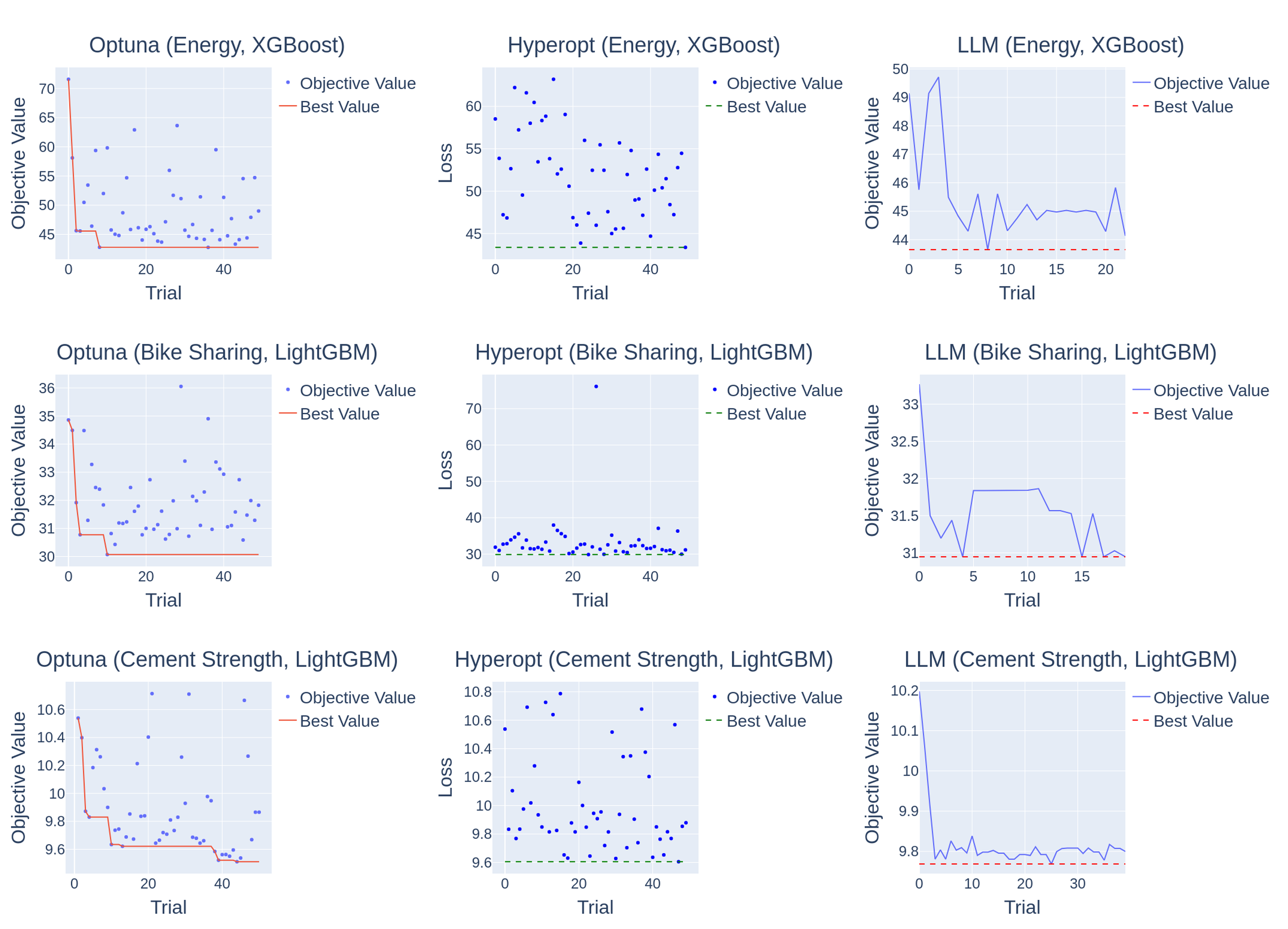}
    \caption{The Figure illustrates the optimization history for Optuna, Hyperopt, and initial LLM HPO strategy with GPT-3.5-Turbo and intelligent\_summary and LLM's early stopping with patience\_15. The top panel represents the energy dataset with the XGBoost model, the middle panel bike sharing dataset with LightGBM, and the bottom panel cement strength dataset with LightGBM.}
    \label{fig:figure_2_init_llm_exp}
\end{figure}

An analysis of the optimization history provides additional insights into the LLM-based approach. For most tasks, the LLM-based method exhibited significantly lower variability in optimization scores than Optuna and Hyperopt. For example, in the energy dataset with XGBoost, the standard deviation of the LLM-based method’s scores was approximately 2.01, compared to 4.92 for Hyperopt and 6.37 for Optuna, as shown in Figure~\ref{fig:figure_2_init_llm_exp}. This lower variability suggests that the LLM method tended to exploit previously learned regions of the hyperparameter space more intensively, leading to narrower score ranges across iterations. This can be interpreted as a sign of overexploitation, where the model focuses on refining a limited set of parameters rather than exploring a broader range of possibilities. In contrast, Optuna and Hyperopt showed broader exploration, reflected in their higher score variability and broader ranges between the worst and best scores.

A notable observation in the LLM-powered experiments was the frequent early stopping of optimization. As illustrated in Figure~\ref{fig:figure_2_init_llm_exp}, unlike Optuna and Hyperopt, which consistently completed 50 iterations, the LLM-based method frequently stopped after around 20, with a maximum of 40 iterations observed in the concrete\_strength dataset with LightGBM.

Despite these challenges, early stopping proved advantageous in some instances, such as in the energy dataset with XGBoost and the M5 dataset with LightGBM, where the LLM-based method achieved superior or comparable results using far fewer iterations. In these cases, the LLM method required, on average, 6 times less computational budget than Optuna and 8.5 times less than Hyperopt while still identifying better or near-optimal parameters (see Table~\ref{tab:table2}). However, while overexploitation worked effectively when the initial parameters were near the global optimum, it limited the search's effectiveness when the initial parameters were in the vicinity of a local optimum. This highlights the need for a better balance between exploration and exploitation, particularly when the starting point is less ideal.

\begin{figure}[!ht]
    \centering
    \includegraphics[width=0.8\textwidth]{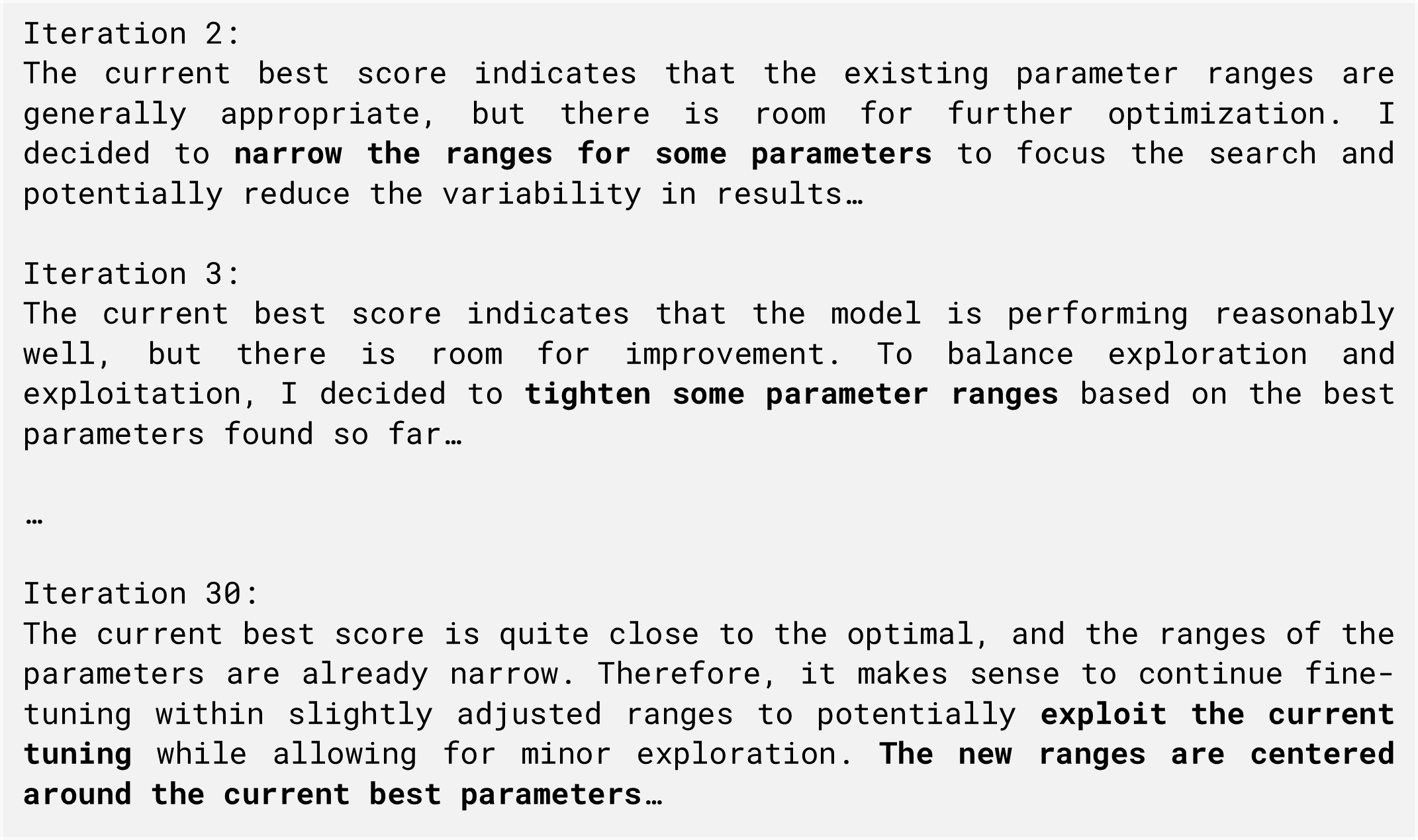}
    \caption{The Figure provides LLM's decisions about parameter ranges for some iterations on the M5 dataset with the LightGBM model starting from the second iteration, as the first iteration is initialization.}
    \label{fig:figure_3_llm_reasoning_output}
\end{figure}

Further analysis revealed that the LLM's decision-making process on search space adjustments significantly shaped the observed optimization behavior. The LLM's reasoning behind these decisions was captured through an additional process called the Reasoning Engine, as shown in Figure~\ref{fig:figure_3_llm_reasoning_output}.

As the analysis progressed, it became clear that early stopping was not solely due to the LLM's optimization behavior. Still, it was also influenced by external factors such as API-related issues and LLM-based history summarization. These technical limitations highlighted the need for more robust memory management and request handling. The decision was made to transition to the LangChain framework to address these technical concerns, which offers automatic memory management and improved error-handling mechanisms, mitigating the issues encountered in the earlier experiments. While this switch was not explicitly aimed at solving the overexploitation problem, LangChain's more stable and controlled environment allowed for further analysis of the implicit effects on the exploration-exploitation trade-off. This transition also allowed for easy experimentation of other LLM models, such as GPT-4o, Gemini, and Claude-Sonnet, to assess whether these models could improve the optimization process. Furthermore, in the next phase of experiments, new sampling strategies specifically designed to address overexploitation will be discussed in detail using LangChain.

\subsection{SLLMBO with Fully LLM-Powered Methods and LangChain}
\label{subsection_4_2}

With the transition to the LangChain framework, the focus shifted to analyzing whether the improved technical setup could positively impact the overall optimization process beyond addressing memory and API-related issues. The experiments in this phase sought to investigate the performance of various LLM models, GPT-3.5-Turbo and GPT-4o, Gemini-1.5-Flash, and Claude-3.5-Sonnet, within LangChain’s more stable environment and to analyze whether these models could further improve the optimization outcomes.

\input{tables/table3} 
\input{tables/table4} 

Regarding parameter initialization, the results remained consistent with earlier findings, where LLM models, especially GPT-3.5-Turbo and Claude-3.5-Sonnet, outperformed traditional random initialization strategies from Optuna and Hyperopt (see Tables~\ref{tab:table3} and~\ref{tab:table4}). All LLM models demonstrated superior initialization compared to the conventional methods in several tasks, further reinforcing the LLM's ability to suggest effective starting points for optimization, particularly in regression tasks. 

When evaluating the overall HPO performance, LangChain-based LLMs provided significant improvements. While the earlier LLM-based method only surpassed Optuna and Hyperopt in two tasks, the LangChain-based strategies outperformed them in eight tasks, with GPT-3.5-Turbo leading in half of these instances (Tables~\ref{tab:table3} and~\ref{tab:table4}). This method demonstrated consistently strong results across different datasets, further solidifying its position as a reliable model for HPO. GPT-4o followed closely, though its average performance was slightly below GPT-3.5-Turbo. Notably, Claude-3.5-Sonnet delivered competitive results in classification tasks, as shown in Table~\ref{tab:table3}, but proved to be the most expensive API usage, making it less efficient in computational cost. Both Tables~\ref{tab:table3} and~\ref{tab:table4} show that Gemini-1.5-Flash, while the fastest and cheapest model, ranked last in performance, with premature stopping in most cases after early promising results. 

\begin{figure}[!htbp] 
    \centering
    \includegraphics[width=\textwidth]{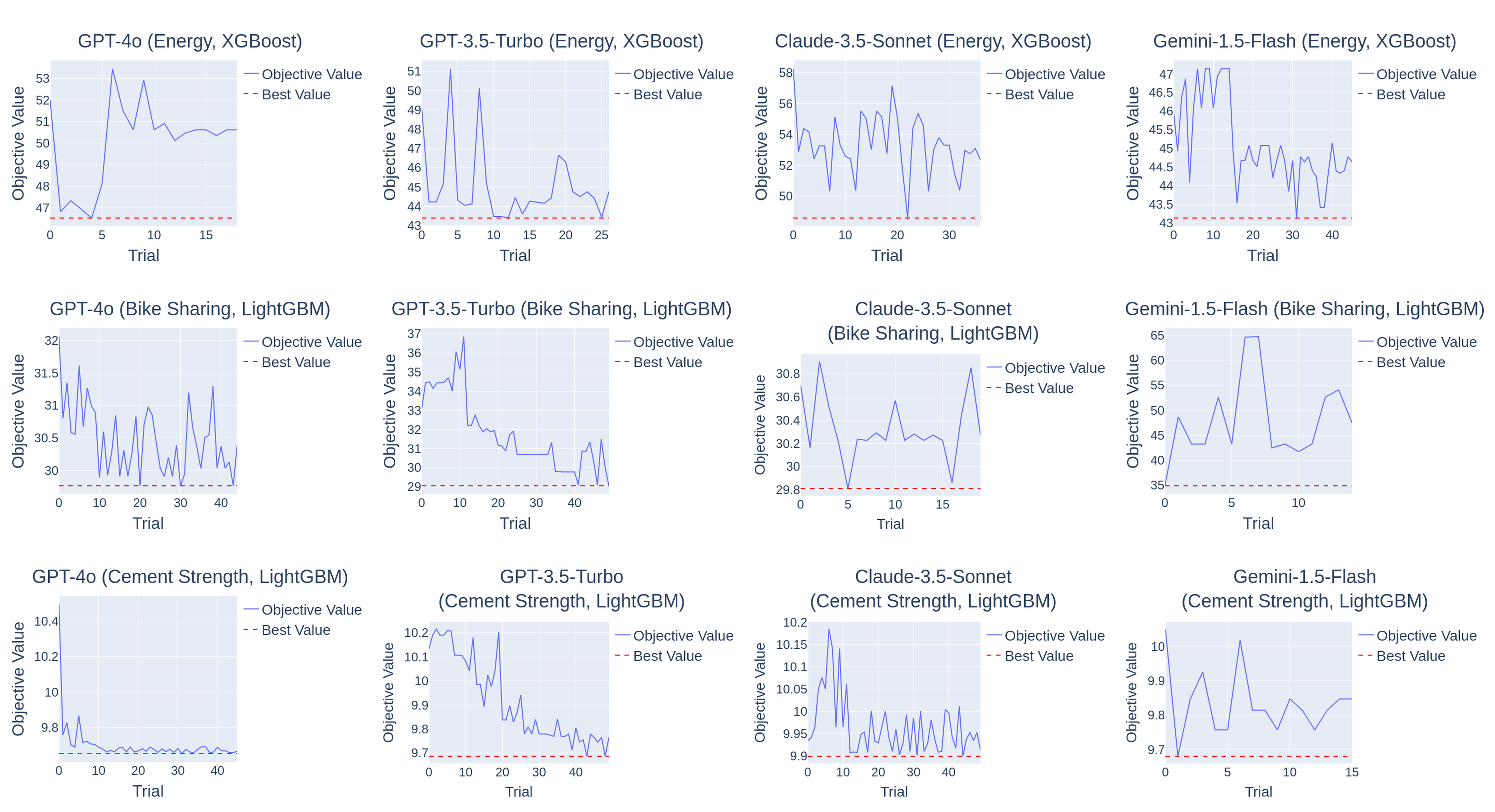}
    \caption{The Figure illustrates the optimization history for LLMs: GPT-4o, GPT-3.5-Turbo, Claude-3.5-Sonnet, and Gemini-1.5-Flash with second LLM strategy, using LangChain to run the LLM APIs, replacing intelligent summary with LangChain's memory buffer. Early Stopping of patience\_15 is used for all LLMs. The top panel represents the energy dataset with the XGBoost model, the middle panel bike sharing dataset with LightGBM, and the bottom panel cement strength dataset with LightGBM.}
    \label{fig:figure4}
\end{figure}

Analyzing the optimization histories reveals that LangChain-based LLMs benefitted from longer runs, improving the balance between exploration and exploitation, particularly in early iterations. GPT-3.5-Turbo, in contrast to its earlier runs with LLM summarization, now consistently reached 50 iterations in nearly half of the tasks (Figure~\ref{fig:figure4}). Although early-stopping issues due to no improvement persisted across all models, this occurred less frequently in GPT-3.5-turbo and GPT-4o compared to the other models. Gemini-1.5-Flash frequently exhibited a pattern of initial solid performance, only to plateau after a few iterations, resulting in premature stopping and consistently lower rankings. Sonnet-3.5, while competitive in some tasks, faced similar issues with overexploitation, where initial gains were not sustained throughout the optimization process. While LangChain significantly improved the technical handling of HPO with LLMs and led to more robust overall performance, the overexploitation problem remains. The next phase discusses the LLM-TPE hybrid sampler, which combines the LLM’s strength in initialization and exploitation with TPE’s proven sampling strategy to achieve a more balanced exploration-exploitation trade-off. This method will be evaluated to determine its effectiveness in addressing the limitations observed in earlier experiments.

\subsection{SLLMBO with LLM-TPE Sampler}

Following the improvements made with LangChain-based LLM methods, this phase addresses the remaining overexploitation issues by integrating the LLM’s initialization and reasoning capabilities with the Tree-structured Parzen Estimator for better sampling. The LLM-TPE hybrid sampler aims to reduce premature early stoppings and explore parameter spaces more thoroughly.

GPT-4o and Gemini-1.5-Flash were selected to achieve this goal. Gemini was the fastest and cheapest model, but it had premature stoppings, so it is included to assess whether the hybrid strategy could improve its performance. GPT-3.5-turbo was excluded due to its consistent success in previous phases, and Sonnet was omitted due to high costs.

\input{tables/table5.tex}
\input{tables/table6.tex}

The GPT-4o-based LLM-TPE sampler was tested with random and LLM initialization strategies to further assess the effectiveness of LLM-based initialization in hyperparameter optimization. The results in Tables~\ref{tab:table5} and~\ref{tab:table6} reaffirmed the strength of LLM-based initialization, as in 13 out of the 14 tasks, GPT-4o's LLM initialization outperformed random initialization in providing a better starting point. However, a more detailed analysis of the final optimized parameters after 50 iterations revealed a notable distinction between regression and classification tasks. In regression tasks, as given in Table~\ref{tab:table6}, the best parameters starting with random initialization consistently resulted in better test scores. In contrast, Table~\ref{tab:table5} shows that, in nearly all classification tasks, starting with GPT-4o initialization led to superior results.

Regarding overall HPO performance, the GPT-4o LLM-TPE sampler in Tables~\ref{tab:table5} and~\ref{tab:table6} consistently outperformed the previous GPT-4o fully LLM sampler (Tables~\ref{tab:table3} and~\ref{tab:table4}) approach in 9 of the 14 tasks, delivering competitive scores in the remaining tasks. Gemini-1.5-Flash, which previously exhibited poor performance and frequent early stoppings, showed significant improvement with the LLM-TPE sampler. The Gemini-based LLM-TPE method outperformed its fully LLM-powered counterpart in 12 tasks, rising from last place in many previous evaluations to a more competitive ranking. Compared to Optuna, Hyperopt, and the fully LLM sampler with GPT-4o and Gemini-1.5-Flash, the LLM-TPE sampler demonstrated superior performance, with GPT-4o consistently ranking in the top one or two positions, followed closely by Gemini-1.5-Flash.

\begin{figure}[!ht]
    \centering
    \includegraphics[width=\textwidth]{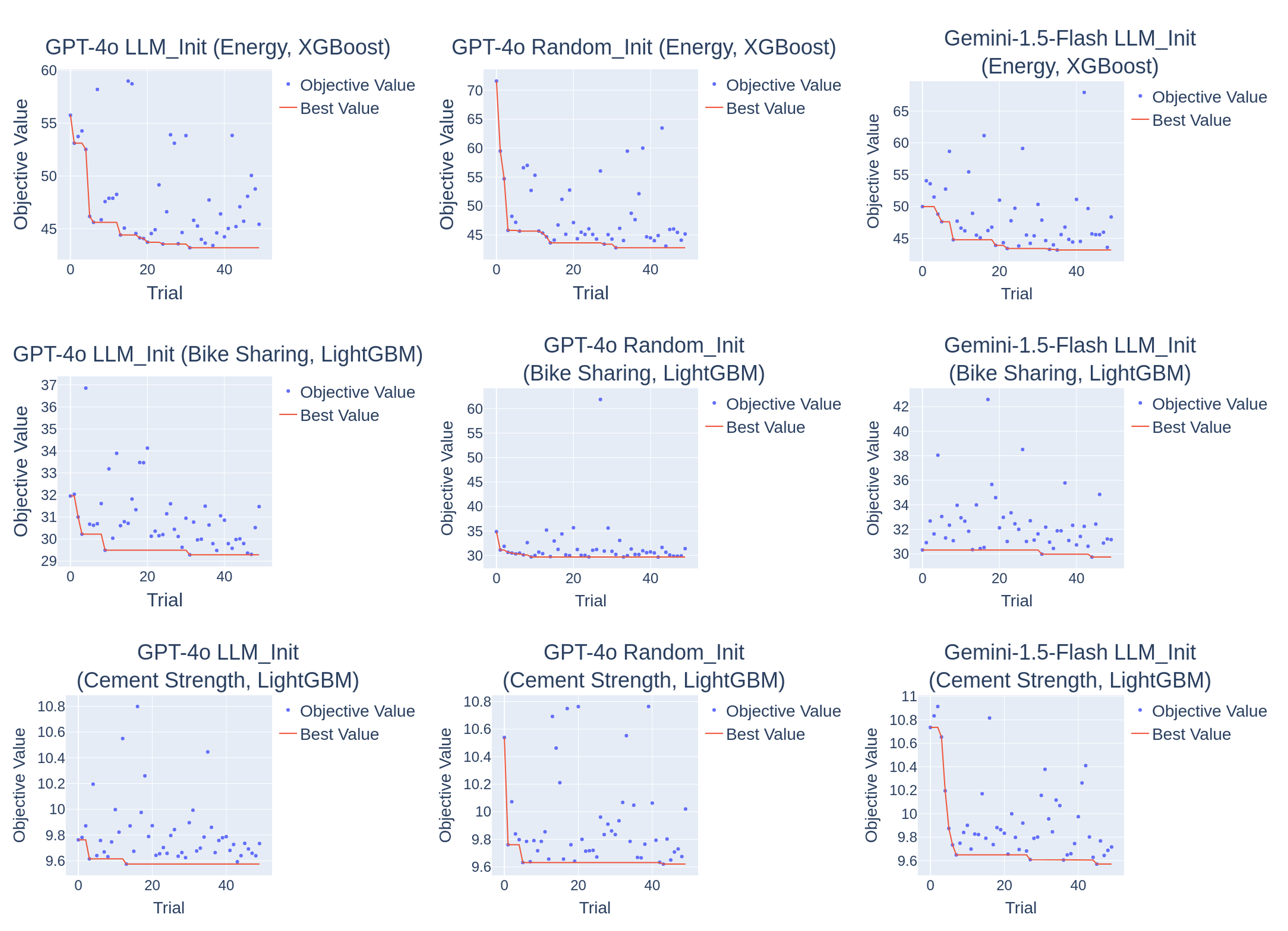}
    \caption{Optimization history plots for LLM-TPE Sampler with GPT-4o with LLM-based initialization, random initialization, and Gemini-1.5-Flash with LLM initialization. Early Stopping of patience\_15 is used for all LLMs. The top panel represents the energy dataset with the XGBoost model, the middle panel bike sharing dataset with LightGBM, and the bottom panel cement strength dataset with LightGBM.}
    \label{fig:figure_5_llm_tpe_exp}
\end{figure}

\input{tables/table7.tex}
\input{tables/table8.tex}

The success of the LLM-TPE sampler can be attributed to its ability to maintain a balanced exploration-exploitation trade-off. The fully LLM-based approach excelled at exploitation, as results from Sections~\ref{subsection_4_1} and~\ref{subsection_4_2} demonstrated. At the same time, the TPE method’s capacity to explore uncertain regions of parameter space is well-established in prior research. By combining these two approaches, the LLM-TPE sampler offered a more holistic and effective optimization strategy, addressing the limitations of using either technique in isolation. This integration significantly reduced the occurrence of premature early stoppings observed in previous methods. As shown from the optimization history plots in Figure~\ref{fig:figure_5_llm_tpe_exp}, the LLM-TPE sampler enabled models to complete 50 iterations with fewer API calls, reducing computational costs. Furthermore, the analysis of optimization scores throughout the trials revealed that the LLM-TPE sampler no longer exhibited the limited exploration and narrow score ranges characteristic of fully LLM-based strategies. Instead, it achieved a more balanced exploration, with score variability comparable to Bayesian Optimization methods. This enhanced exploration facilitated the discovery of more optimal parameter configurations, contributing to the improved performance observed across multiple tasks.

Early stopping (ES) with patience of 5 is also tested to evaluate the effectiveness of the LLM-TPE sampler in budget-constrained scenarios. This experiment focused on the Gemini-1.5-Flash LLM-TPE sampler, comparing its performance with and without early stopping and against Optuna to determine how limiting the number of iterations influences hyperparameter optimization outcomes. While early stopping reduced computational costs across tasks, it also affected performance (Tables~\ref{tab:table7},~\ref{tab:table8}). Gemini with ES outperformed Optuna with ES in 6 tasks, yet both showed decreased performance compared to their non-ES counterparts. Optuna benefitted from early stopping in specific tasks, while Gemini exhibited a consistent performance drop.

\begin{figure}[!ht]
    \centering
    \includegraphics[width=\textwidth]{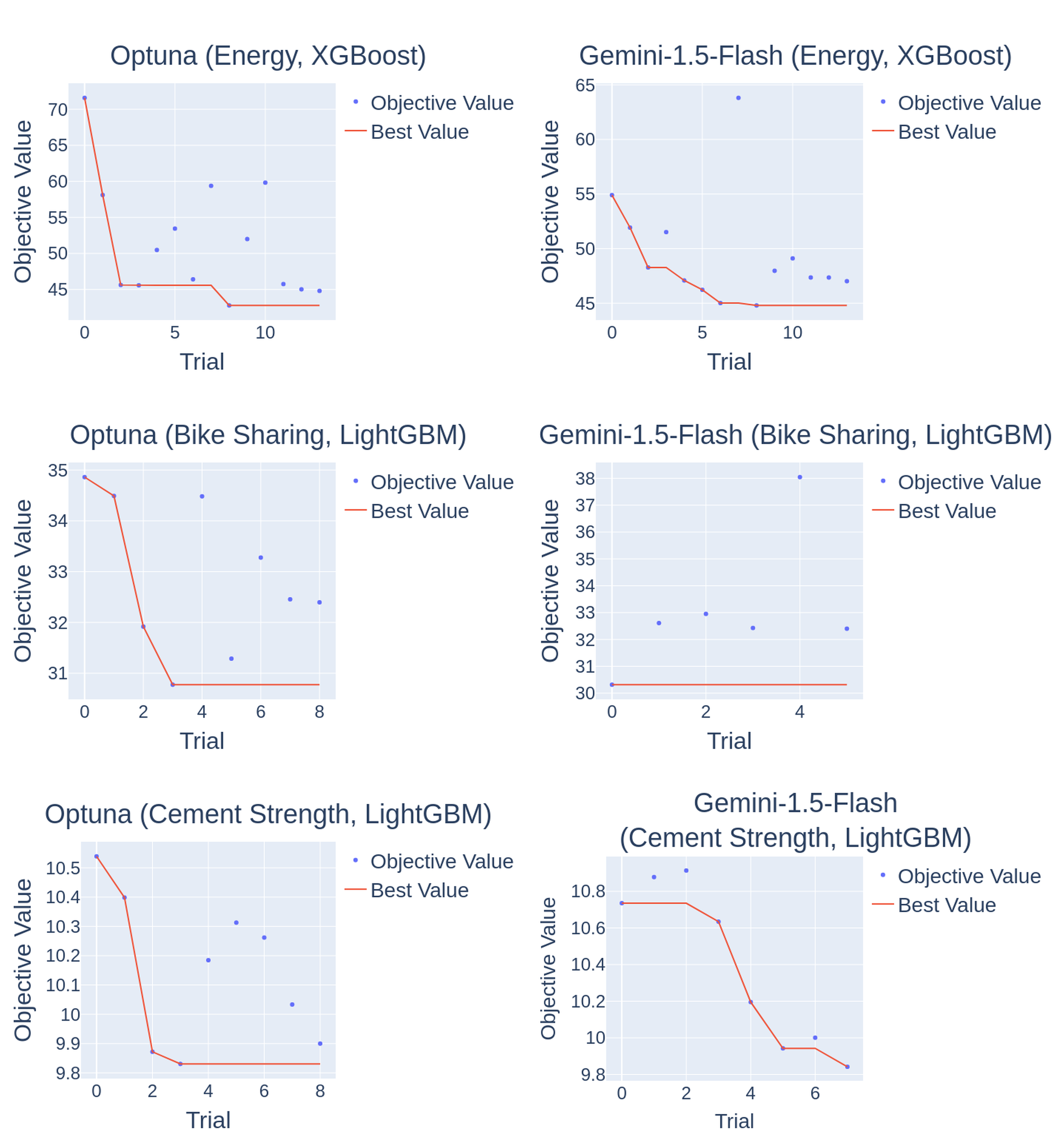}
    \caption{Optimization history plots for Optuna and LLM-TPE Sampler with Gemini-1.5-Flash with LLM initialization, both having Early Stopping of patience\_5. The top panel represents the energy dataset with the XGBoost model, the middle panel bike sharing dataset with LightGBM, and the bottom panel cement strength dataset with LightGBM.}
    \label{fig:figure_6_llm_tpe_es_exp}
\end{figure}

Figure~\ref{fig:figure_6_llm_tpe_es_exp} also reveals that both Optuna and the LLM-TPE sampler run around ten iterations with strict regularization. These findings suggest that LLM-TPE-based methods may require more iterations than Bayesian Optimization methods to fully exploit their alternating sampling strategies and achieve optimal results.

%% file: tables/table1.tex
\begin{table}[!ht]
    \centering
    \small 
    \renewcommand{\arraystretch}{1.5} 
    \setlength{\tabcolsep}{6pt} 
    \resizebox{\textwidth}{!}{%
        \begin{tabular}{@{}lllS[table-format=2.2]S[table-format=1.2]S[table-format=2.2]S[table-format=4.0]@{}}
        \toprule
        \textbf{Dataset} & \textbf{Model} & \textbf{Optimizer} & {\makecell{\textbf{Initial Score} \\ \textbf{(CV Avg)}}} & {\makecell{\textbf{Optimized Score} \\ \textbf{(CV Avg)}}} & {\textbf{Test Score}} & {\textbf{Runtime (s)}} \\
        \midrule
        
        \multirow{6}{*}{Gas Drift} & \multirow{3}{*}{LightGBM} 
        & Optuna & 0.7855 & 0.8187 & 0.8542 & 2620 \\
        & & Hyperopt & 0.7898 & 0.8191 & 0.8417 & 702 \\
        & & \textbf{Ours (LLM)} & 0.8162 & 0.8218 & 0.8427 & 1680 \\
        \cmidrule{2-7}
        & \multirow{3}{*}{XGBoost}
        & Optuna & 0.8002 & 0.8138 & 0.8311 & 2019 \\
        & & Hyperopt & 0.7582 & 0.8076 & 0.8224 & 69 \\
        & & \textbf{Ours (LLM)} & 0.7546 & 0.7678 & 0.7929 & 67 \\
        
        \midrule
        
        \multirow{6}{*}{Cover Type} & \multirow{3}{*}{LightGBM}
        & Optuna & 0.7043 & 0.7115 & 0.7111 & 1093 \\
        & & Hyperopt & 0.7046 & 0.7099 & 0.7056 & 842 \\
        & & \textbf{Ours (LLM)} & 0.7082 & 0.7126 & 0.7069 & 189 \\
        \cmidrule{2-7}
        & \multirow{3}{*}{XGBoost}
        & Optuna & 0.6871 & 0.7140 & 0.7035 & 1033 \\
        & & Hyperopt & 0.7090 & 0.7154 & 0.7028 & 853 \\
        & & \textbf{Ours (LLM)} & 0.7093 & 0.7106 & 0.7032 & 134 \\
        
        \midrule
        
        \multirow{6}{*}{Adult Census} & \multirow{3}{*}{LightGBM}
        & Optuna & 0.7789 & 0.7822 & 0.7824 & 140 \\
        & & Hyperopt & 0.8117 & 0.7823 & 0.7821 & 96 \\
        & & \textbf{Ours (LLM)} & 0.7769 & 0.7813 & 0.7787 & 138 \\
        \cmidrule{2-7}
        & \multirow{3}{*}{XGBoost}
        & Optuna & 0.7725 & 0.7823 & 0.7856 & 112 \\
        & & Hyperopt & 0.7788 & 0.7826 & 0.7826 & 151 \\
        & & \textbf{Ours (LLM)} & 0.7773 & 0.7809 & 0.7812 & 123 \\
        \bottomrule
        \end{tabular}%
    }
\caption{Comparison of Optuna (w/o patience), Hyperopt (w/o patience), and initial LLM HPO strategy with GPT-3.5-Turbo and intelligent\_summary for \textbf{classification} tasks with LightGBM and XGBoost models with f1-weighted as the evaluation metric. \textbf{LLM} strategy corresponds to the reference: "llm\_sampler+intelligent\_summary+gpt-3.5-turbo+patience\_15" defined in the Methodology~\ref{methodology:llm_strategy1}.}
\label{tab:table1}
\end{table}

%% file: tables/table2.tex
\begin{table}[!ht]
    \centering
    \small 
    \renewcommand{\arraystretch}{1.5} 
    \setlength{\tabcolsep}{6pt} 
    \resizebox{\textwidth}{!}{%
        \begin{tabular}{@{}lllS[table-format=2.2]S[table-format=1.2]S[table-format=2.2]S[table-format=4.0]@{}}
        \textbf{Dataset} & \textbf{Model} & \textbf{Optimizer} & {\makecell{\textbf{Initial Score} \\ \textbf{(CV Avg)}}} & {\makecell{\textbf{Optimized Score} \\ \textbf{(CV Avg)}}} & {\textbf{Test Score}} & {\textbf{Runtime (s)}} \\
        \midrule
        
        \multirow{6}{*}{Bike Sharing} & \multirow{3}{*}{LightGBM} 
        & Optuna & 34.9221 & 30.0695 & 65.5113 & 394 \\
        & & Hyperopt & 32.5112 & 29.8674 & 64.2410 & 284 \\
        & & \textbf{Ours (LLM)} & 33.5116 & 30.9468 & 66.6167 & 94 \\
        \cmidrule{2-7}
        & \multirow{3}{*}{XGBoost}
        & Optuna & 38.8547 & 29.4815 & 58.9873 & 585 \\
        & & Hyperopt & 37.3286 & 29.9668 & 61.1801 & 380 \\
        & & \textbf{Ours (LLM)} & 33.8210 & 30.8633 & 63.5643 & 63 \\
        
        \midrule
        
        \multirow{6}{*}{Concrete Strength} & \multirow{3}{*}{LightGBM}
        & Optuna & 10.5745 & 9.5110 & 9.8793 & 51 \\
        & & Hyperopt & 10.5624 & 9.6051 & 10.0190 & 32 \\
        & & \textbf{Ours (LLM)} & 10.2010 & 9.7688 & 9.9930 & 122 \\
        \cmidrule{2-7}
        & \multirow{3}{*}{XGBoost}
        & Optuna & 11.7358 & 9.6385 & 9.9103 & 92 \\
        & & Hyperopt & 11.2125 & 9.5924 & 9.8731 & 137 \\
        & & \textbf{Ours (LLM)} & 10.2235 & 9.6220 & 9.8794 & 58 \\
        
        \midrule
        
        \multirow{6}{*}{Energy} & \multirow{3}{*}{LightGBM}
        & Optuna & 55.2416 & 42.5948 & 125.5641 & 265 \\
        & & Hyperopt & 50.8772 & 43.2999 & 99.4970 & 303 \\
        & & \textbf{Ours (LLM)} & 47.4421 & 43.4206 & 157.1979 & 64 \\
        \cmidrule{2-7}
        & \multirow{3}{*}{XGBoost}
        & Optuna & 72.3590 & 42.7662 & 160.7597 & 381 \\
        & & Hyperopt & 57.5863 & 43.3675 & 141.2623 & 796 \\
        & & \textbf{Ours (LLM)} & 49.1075 & 43.6496 & 136.7344 & 119 \\
        
        \midrule

        \multirow{6}{*}{M5} & \multirow{3}{*}{LightGBM}
        & Optuna & 6.5749 & 6.1575 & 5.9046 & 585 \\
        & & Hyperopt & 6.3228 & 6.1460 & 5.9169 & 610 \\
        & & \textbf{Ours (LLM)} & 6.2814 & 6.1245 & 5.6220 & 617 \\
        \cmidrule{2-7}
        & \multirow{3}{*}{XGBoost}
        & Optuna & 7.2866 & 6.0656 & 5.9303 & 1434 \\
        & & Hyperopt & 6.6237 & 6.0534 & 5.6242 & 1622 \\
        & & \textbf{Ours (LLM)} & 6.2553 & 6.2244 & 5.8178 & 161 \\
        \bottomrule
        \end{tabular}%
    }
\caption{Comparison of Optuna (w/o patience), Hyperopt (w/o patience), and initial LLM HPO strategy with GPT-3.5-Turbo and intelligent\_summary for \textbf{regression} tasks with LightGBM and XGBoost models. \textbf{LLM} optimizer corresponds to the reference: "llm\_sampler+intelligent\_summary+gpt-3.5-turbo+patience\_15" defined in the Methodology~\ref{methodology:llm_strategy1}.}
\label{tab:table2}
\end{table}

%% file: tables/table3.tex
\begin{table}[!ht]
    \centering
    \small 
    \renewcommand{\arraystretch}{1.5} 
    \setlength{\tabcolsep}{6pt} 
    \resizebox{\textwidth}{!}{%
        \begin{tabular}{@{}lllS[table-format=2.2]S[table-format=1.2]S[table-format=2.2]S[table-format=4.0]@{}}
        \textbf{Dataset} & \textbf{Model} & 
        {\makecell{\textbf{Optimizer}
        \\ \textbf{(LangChain Memory Buffer)}}}
        & {\makecell{\textbf{Initial Score} \\ \textbf{(CV Avg)}}} & {\makecell{\textbf{Optimized Score} \\ \textbf{(CV Avg)}}} & {\textbf{Test Score}} & {\textbf{Runtime (s)}} \\
        \midrule
        
        \multirow{8}{*}{Gas Drift} & \multirow{4}{*}{LightGBM} 
        & Gpt-4o & 0.7930 & 0.8198 & 0.8447 & 443 \\
        & & Gpt-3.5-Turbo & 0.7745 & 0.8253 & 0.8527 & 1454 \\
        & & Claude-3.5-Sonnet & 0.7562 & 0.8139 & 0.8301 & 543 \\
        & & Gemini-1.5-Flash & 0.8004 & 0.8004 & 0.8192 & 84 \\
        \cmidrule{2-7}
        & \multirow{4}{*}{XGBoost}
        & Gpt-4o & 0.7760 & 0.8087 & 0.8197 & 742 \\
        & & Gpt-3.5-Turbo & 0.7570 & 0.8183 & 0.8392 & 539 \\
        & & Claude-3.5-Sonnet & 0.7758 & 0.7878 & 0.8075 & 502 \\
        & & Gemini-1.5-Flash & 0.7758 & 0.7758 & 0.7913 & 131 \\
        
        \midrule
        
        \multirow{8}{*}{Cover Type} & \multirow{4}{*}{LightGBM}
        & Gpt-4o & 0.6975 & 0.7014 & 0.6910 & 1433 \\
        & & Gpt-3.5-Turbo & 0.7092 & 0.7104 & 0.7085 & 173 \\
        & & Claude-3.5-Sonnet & 0.7080 & 0.7123 & 0.7086 & 525 \\
        & & Gemini-1.5-Flash & 0.7052 & 0.7115 & 0.7022 & 180 \\
        \cmidrule{2-7}
        & \multirow{4}{*}{XGBoost}
        & Gpt-4o & 0.7128 & 0.7135 & 0.7037 & 222 \\
        & & Gpt-3.5-Turbo & 0.7093 & 0.7140 & 0.7025 & 674 \\
        & & Claude-3.5-Sonnet & 0.7128 & 0.7128 & 0.7040 & 238 \\
        & & Gemini-1.5-Flash & 0.7128 & 0.7128 & 0.7040 & 140 \\
        
        \midrule
        
        \multirow{8}{*}{Adult Census} & \multirow{4}{*}{LightGBM}
        & Gpt-4o & 0.7780 & 0.7789 & 0.7767 & 186 \\
        & & Gpt-3.5-Turbo & 0.7725 & 0.7783 & 0.7833 & 46 \\
        & & Claude-3.5-Sonnet & 0.7784 & 0.7808 & 0.7808 & 125 \\
        & & Gemini-1.5-Flash & 0.7776 & 0.7811 & 0.7823 & 122 \\
        \cmidrule{2-7}
        & \multirow{4}{*}{XGBoost}
        & Gpt-4o & 0.7804 & 0.7820 & 0.7809 & 142 \\
        & & Gpt-3.5-Turbo & 0.7802 & 0.7817 & 0.7839 & 44 \\
        & & Claude-3.5-Sonnet & 0.7762 & 0.7768 & 0.7803 & 175 \\
        & & Gemini-1.5-Flash & 0.7809 & 0.7854 & 0.7905 & 116 \\
        \bottomrule
        \end{tabular}%
    }
\caption{This table represents the results of the second LLM strategy, which replaces intelligent summary in previous experimentation with LangChain memory buffer, and compares the following LLMs\: GPT-4o, GPT-3.5-Turbo, Claude-3.5-Sonnet, and Gemini-1.5-Flash for \textbf{classification} tasks with LightGBM and XGBoost models. This strategy corresponds to the reference: "llm\_sampler+langchain+\{\textit{optimizer}\}+patience\_15" defined in the Methodology~\ref{methodology:llm_strategy2}.}
\label{tab:table3}
\end{table}

%% file: tables/table4.tex
\begin{table}[!ht]
    \centering
    \small 
    \renewcommand{\arraystretch}{1.5} 
    \setlength{\tabcolsep}{6pt} 
    \resizebox{\textwidth}{!}{%
        \begin{tabular}{@{}lllS[table-format=2.2]S[table-format=1.2]S[table-format=2.2]S[table-format=4.0]@{}}
        \textbf{Dataset} & \textbf{Model} & {\makecell{\textbf{Optimizer}
        \\ \textbf{(LangChain Memory Buffer)}}} & {\makecell{\textbf{Initial Score} \\ \textbf{(CV Avg)}}} & {\makecell{\textbf{Optimized Score} \\ \textbf{(CV Avg)}}} & {\textbf{Test Score}} & {\textbf{Runtime (s)}} \\
        \midrule
        
        \multirow{8}{*}{Bike Sharing} & \multirow{4}{*}{LightGBM} 
        & Gpt-4o & 32.1523 & 29.7614 & 57.1713 & 470 \\
        & & Gpt-3.5-Turbo & 33.0547 & 29.0524 & 63.8804 & 126 \\
        & & Claude-3.5-Sonnet & 30.7225 & 29.8112 & 61.2840 & 145 \\
        & & Gemini-1.5-Flash & 35.0186 & 34.8946 & 67.1382 & 54 \\
        \cmidrule{2-7}
        & \multirow{4}{*}{XGBoost}
        & Gpt-4o & 32.2341 & 29.1302 & 59.3126 & 128 \\
        & & Gpt-3.5-Turbo & 31.7440 & 30.4025 & 58.0684 & 272 \\
        & & Claude-3.5-Sonnet & 32.3194 & 31.7193 & 59.7128 & 235 \\
        & & Gemini-1.5-Flash & 31.4992 & 31.4117 & 64.7856 & 74 \\
        
        \midrule
        
        \multirow{8}{*}{Concrete Strength} & \multirow{4}{*}{LightGBM}
        & Gpt-4o & 10.5153 & 9.6535 & 9.9843 & 153 \\
        & & Gpt-3.5-Turbo & 10.1347 & 9.6861 & 10.0641 & 101 \\
        & & Claude-3.5-Sonnet & 9.9385 & 9.8995 & 10.0505 & 287 \\
        & & Gemini-1.5-Flash & 10.0510 & 9.6807 & 10.0888 & 47 \\
        \cmidrule{2-7}
        & \multirow{4}{*}{XGBoost}
        & Gpt-4o & 10.5890 & 10.3823 & 10.0358 & 147 \\
        & & Gpt-3.5-Turbo & 10.4652 & 9.9616 & 9.9059 & 35 \\
        & & Claude-3.5-Sonnet & 11.0380 & 10.8452 & 10.1716 & 142 \\
        & & Gemini-1.5-Flash & 10.4225 & 9.9162 & 9.9833 & 69 \\
        
        \midrule
        
        \multirow{8}{*}{Energy} & \multirow{4}{*}{LightGBM}
        & Gpt-4o & 43.3435 & 43.3435 & 135.4801 & 122 \\
        & & Gpt-3.5-Turbo & 43.8546 & 42.8751 & 145.0990 & 60 \\
        & & Claude-3.5-Sonnet & 45.3893 & 42.7568 & 143.6961 & 375 \\
        & & Gemini-1.5-Flash & 45.9150 & 44.7903 & 116.5710 & 32 \\
        \cmidrule{2-7}
        & \multirow{4}{*}{XGBoost}
        & Gpt-4o & 51.9850 & 46.5245 & 163.9273 & 246 \\
        & & Gpt-3.5-Turbo & 49.2038 & 43.3944 & 157.4147 & 139 \\
        & & Claude-3.5-Sonnet & 58.4267 & 48.5841 & 198.2501 & 734 \\
        & & Gemini-1.5-Flash & 45.9780 & 43.1292 & 164.9638 & 346 \\
        
        \midrule

        \multirow{8}{*}{M5} & \multirow{4}{*}{LightGBM}
        & Gpt-4o & 6.4130 & 6.2537 & 5.5975 & 304 \\
        & & Gpt-3.5-Turbo & 6.2394 & 6.1088 & 5.7099 & 450 \\
        & & Claude-3.5-Sonnet & 6.2980 & 6.1581 & 5.6566 & 979 \\
        & & Gemini-1.5-Flash & 6.4423 & 6.4423 & 5.7213 & 93 \\
        \cmidrule{2-7}
        & \multirow{4}{*}{XGBoost}
        & Gpt-4o & 6.2582 & 6.1391 & 5.6706 & 584 \\
        & & Gpt-3.5-Turbo & 6.2950 & 6.0747 & 5.8179 & 2408 \\
        & & Claude-3.5-Sonnet & 6.1725 & 6.0525 & 5.8003 & 769 \\
        & & Gemini-1.5-Flash & 6.1714 & 6.0264 & 5.7672 & 1439 \\
        \bottomrule
        \end{tabular}%
    }
\caption{This table represents the results of the second LLM strategy, which replaces intelligent summary in previous experimentation with LangChain memory buffer, and compares the following LLMs\: GPT-4o, GPT-3.5-Turbo, Claude-3.5-Sonnet, and Gemini-1.5-Flash for \textbf{regression} tasks with LightGBM and XGBoost models. This strategy corresponds to the reference: "llm\_sampler+langchain+\{\textit{optimizer}\}+patience\_15" defined in the Methodology~\ref{methodology:llm_strategy2}.}
\label{tab:table4}
\end{table}

%% file: tables/table5.tex
\begin{table}[!ht]
    \centering
    \small 
    \renewcommand{\arraystretch}{1.5} 
    \setlength{\tabcolsep}{6pt} 
    \resizebox{\textwidth}{!}{%
        \begin{tabular}{@{}lllS[table-format=2.2]S[table-format=1.2]S[table-format=2.2]S[table-format=4.0]@{}}
        \textbf{Dataset} & \textbf{Model} & {\makecell{\textbf{Optimizer}
        \\ \textbf{(LLM-TPE Sampler)}}} & {\makecell{\textbf{Initial Score} \\ \textbf{(CV Avg)}}} & {\makecell{\textbf{Optimized Score} \\ \textbf{(CV Avg)}}} & {\textbf{Test Score}} & {\textbf{Runtime (s)}} \\
        \midrule
        
        \multirow{6}{*}{Gas Drift} & \multirow{3}{*}{LightGBM} 
        & GPT-4o LLM\_Init & 0.7951 & 0.8236 & 0.8478 & 887 \\
        & & GPT-4o Random\_Init & 0.7864 & 0.8221 & 0.8443 & 1914 \\
        & & Gemini-1.5-Flash LLM\_Init & 0.8160 & 0.8229 & 0.8465 & 1208 \\
        \cmidrule{2-7}
        & \multirow{3}{*}{XGBoost}
        & GPT-4o LLM\_Init & 0.7394 & 0.8180 & 0.8388 & 1013 \\
        & & GPT-4o Random\_Init & 0.8010 & 0.8167 & 0.8400 & 836 \\
        & & Gemini-1.5-Flash LLM\_Init & 0.7786 & 0.8210 & 0.8354 & 1142 \\
        
        \midrule
        
        \multirow{6}{*}{Cover Type} & \multirow{3}{*}{LightGBM}
        & GPT-4o LLM\_Init & 0.7005 & 0.7110 & 0.7056 & 1427 \\
        & & GPT-4o Random\_Init & 0.7043 & 0.7083 & 0.7029 & 1759 \\
        & & Gemini-1.5-Flash LLM\_Init & 0.6963 & 0.7114 & 0.7085 & 2734 \\
        \cmidrule{2-7}
        & \multirow{3}{*}{XGBoost}
        & GPT-4o LLM\_Init & 0.7065 & 0.7166 & 0.7099 & 999 \\
        & & GPT-4o Random\_Init & 0.6869 & 0.7158 & 0.7040 & 780 \\
        & & Gemini-1.5-Flash LLM\_Init & 0.7093 & 0.7159 & 0.7042 & 1661 \\
        
        \midrule
        
        \multirow{6}{*}{Adult Census} & \multirow{3}{*}{LightGBM}
        & GPT-4o LLM\_Init & 0.7798 & 0.7823 & 0.7841 & 290 \\
        & & GPT-4o Random\_Init & 0.7792 & 0.7826 & 0.7821 & 225 \\
        & & Gemini-1.5-Flash LLM\_Init & 0.7753 & 0.7825 & 0.7830 & 501 \\
        \cmidrule{2-7}
        & \multirow{3}{*}{XGBoost}
        & GPT-4o LLM\_Init & 0.7817 & 0.7824 & 0.7834 & 150 \\
        & & GPT-4o Random\_Init & 0.7728 & 0.7823 & 0.7837 & 180 \\
        & & Gemini-1.5-Flash LLM\_Init & 0.7810 & 0.7822 & 0.7849 & 323 \\
        \bottomrule
        \end{tabular}%
    }
\caption{This table represents the results of a novel hybrid \textbf{LLM-TPE Sampler} with LLMs GPT-4o and Gemini-1.5-Flash with LLM-based initialization, also GPT-4o with random initialization for \textbf{classification} tasks with LightGBM and XGBoost models. This strategy corresponds to the reference\: \\ "llm-tpe\_sampler+\{\textit{optimizer}\}+\{\textit{initializer}\}+patience\_15" defined in the Methodology~\ref{methodology:llm_strategy3}.}
\label{tab:table5}
\end{table}

%% file: tables/table6.tex
\begin{table}[!ht]
    \centering
    \small 
    \renewcommand{\arraystretch}{1.5} 
    \setlength{\tabcolsep}{6pt} 
    \resizebox{\textwidth}{!}{%
        \begin{tabular}{@{}lllS[table-format=2.2]S[table-format=1.2]S[table-format=2.2]S[table-format=4.0]@{}}
        \textbf{Dataset} & \textbf{Model} & {\makecell{\textbf{Optimizer}
        \\ \textbf{(LLM-TPE Sampler)}}} & {\makecell{\textbf{Initial Score} \\ \textbf{(CV Avg)}}} & {\makecell{\textbf{Optimized Score} \\ \textbf{(CV Avg)}}} & {\textbf{Test Score}} & {\textbf{Runtime (s)}} \\
        \midrule
        
        \multirow{6}{*}{Bike Sharing} & \multirow{3}{*}{LightGBM} 
        & GPT-4o LLM\_Init & 31.9820 & 29.2870 & 62.1119 & 534 \\
        & & GPT-4o Random\_Init & 35.0012 & 29.6673 & 66.5668 & 683 \\
        & & Gemini-1.5-Flash LLM\_Init & 30.2445 & 29.7375 & 63.0409 & 697 \\
        \cmidrule{2-7}
        & \multirow{3}{*}{XGBoost}
        & GPT-4o LLM\_Init & 30.2007 & 29.2030 & 61.6200 & 339 \\
        & & GPT-4o Random\_Init & 39.1520 & 29.2148 & 60.6247 & 457 \\
        & & Gemini-1.5-Flash LLM\_Init & 32.9483 & 29.1444 & 58.4914 & 599 \\
        
        \midrule
        
        \multirow{6}{*}{Concrete Strength} & \multirow{3}{*}{LightGBM}
        & GPT-4o LLM\_Init & 9.7862 & 9.5747 & 9.8915 & 138 \\
        & & GPT-4o Random\_Init & 10.5783 & 9.6200 & 9.8774 & 156 \\
        & & Gemini-1.5-Flash LLM\_Init & 10.7241 & 9.5704 & 9.8064 & 464 \\
        \cmidrule{2-7}
        & \multirow{3}{*}{XGBoost}
        & GPT-4o LLM\_Init & 11.1031 & 9.7246 & 9.8932 & 155 \\
        & & GPT-4o Random\_Init & 11.7240 & 9.5558 & 9.9134 & 133 \\
        & & Gemini-1.5-Flash LLM\_Init & 10.8026 & 9.5195 & 9.9109 & 323 \\
        
        \midrule
        
        \multirow{6}{*}{Energy} & \multirow{3}{*}{LightGBM}
        & GPT-4o LLM\_Init & 45.3214 & 42.8380 & 116.0518 & 271 \\
        & & GPT-4o Random\_Init & 55.3281 & 41.8082 & 110.3268 & 596 \\
        & & Gemini-1.5-Flash LLM\_Init & 50.9519 & 42.2599 & 112.1581 & 633 \\
        \cmidrule{2-7}
        & \multirow{3}{*}{XGBoost}
        & GPT-4o LLM\_Init & 56.0175 & 43.2021 & 180.3868 & 411 \\
        & & GPT-4o Random\_Init & 73.1084 & 42.7945 & 162.0911 & 592 \\
        & & Gemini-1.5-Flash LLM\_Init & 50.0013 & 43.1632 & 140.5762 & 637 \\
        
        \midrule

        \multirow{6}{*}{M5} & \multirow{3}{*}{LightGBM}
        & GPT-4o LLM\_Init & 6.2406 & 6.1426 & 5.8724 & 826 \\
        & & GPT-4o Random\_Init & 6.5725 & 6.1136 & 5.6316 & 1163 \\
        & & Gemini-1.5-Flash LLM\_Init & 6.2327 & 6.1457 & 5.6703 & 1685 \\
        \cmidrule{2-7}
        & \multirow{3}{*}{XGBoost}
        & GPT-4o LLM\_Init & 6.4007 & 6.0147 & 5.7366 & 2005 \\
        & & GPT-4o Random\_Init & 7.0974 & 5.9907 & 5.7014 & 2422 \\
        & & Gemini-1.5-Flash LLM\_Init & 6.2250 & 6.0215 & 5.7688 & 3211 \\
        \bottomrule
        \end{tabular}%
    }
\caption{This table represents the results of a novel hybrid \textbf{LLM-TPE Sampler} with LLMs GPT-4o and Gemini-1.5-Flash with LLM-based initialization, also GPT-4o with random initialization for \textbf{regression} tasks with LightGBM and XGBoost models. This strategy corresponds to the reference\: \\ "llm-tpe\_sampler+\{\textit{optimizer}\}+\{\textit{initializer}\}+patience\_15" defined in the Methodology~\ref{methodology:llm_strategy3}.}
\label{tab:table6}
\end{table}

%% file: tables/table7.tex
\begin{table}[!ht]
    \centering
    \small 
    \renewcommand{\arraystretch}{1.5} 
    \setlength{\tabcolsep}{6pt} 
    \resizebox{\textwidth}{!}{%
        \begin{tabular}{@{}lllS[table-format=2.2]S[table-format=1.2]S[table-format=2.2]S[table-format=4.0]@{}}
        \textbf{Dataset} & \textbf{Model} & {\makecell{\textbf{Optimizer}
        \\ \textbf{(LLM-TPE Sampler)}}} & {\makecell{\textbf{Initial Score} \\ \textbf{(CV Avg)}}} & {\makecell{\textbf{Optimized Score} \\ \textbf{(CV Avg)}}} & {\textbf{Test Score}} & {\textbf{Runtime (s)}} \\
        \midrule
        
        \multirow{4}{*}{Gas Drift} & \multirow{2}{*}{LightGBM} 
        & Optuna & 0.7875 & 0.8083 & 0.8304 & 371 \\
        & & \textbf{Ours (Gemini-1.5-Flash)} & 0.8148 & 0.8176 & 0.8421 & 237 \\
        \cmidrule{2-7}
        & \multirow{2}{*}{XGBoost}
        & Optuna & 0.8009 & 0.8009 & 0.8224 & 69 \\
        & & \textbf{Ours (Gemini-1.5-Flash)} & 0.7795 & 0.8185 & 0.8393 & 214 \\
        
        \midrule
        
        \multirow{4}{*}{Cover Type} & \multirow{2}{*}{LightGBM}
        & Optuna & 0.7044 & 0.7097 & 0.7050 & 225 \\
        & & \textbf{Ours (Gemini-1.5-Flash)} & 0.7013 & 0.7057 & 0.6955 & 519 \\
        \cmidrule{2-7}
        & \multirow{2}{*}{XGBoost}
        & Optuna & 0.6852 & 0.7102 & 0.7063 & 109 \\
        & & \textbf{Ours (Gemini-1.5-Flash)} & 0.7071 & 0.7155 & 0.7034 & 328 \\
        
        \midrule
        
        \multirow{4}{*}{Adult Census} & \multirow{2}{*}{LightGBM}
        & Optuna & 0.7792 & 0.7821 & 0.7832 & 53 \\
        & & \textbf{Ours (Gemini-1.5-Flash)} & 0.7760 & 0.7811 & 0.7818 & 112 \\
        \cmidrule{2-7}
        & \multirow{2}{*}{XGBoost}
        & Optuna & 0.7728 & 0.7817 & 0.7813 & 42 \\
        & & \textbf{Ours (Gemini-1.5-Flash)} & 0.7807 & 0.7815 & 0.7832 & 77 \\
        \bottomrule
        \end{tabular}%
    }
\caption{This table compares LLM-TPE Sampler with Gemini-1.5-Flash with LLM initialization and Optuna in budget-constrained settings for \textbf{classification} tasks with LightGBM and XGBoost models. For each setup, an early stopping of 5 iterations is used for regularization. LLM strategy corresponds to the reference: "llm-tpe\_sampler+gemini-1.5-flash+llm\_init+patience\_5" defined in the Methodology~\ref{methodology:llm_strategy3}.}
\label{tab:table7}
\end{table}

%% file: tables/table8.tex
\begin{table}[!ht]
    \centering
    \small 
    \renewcommand{\arraystretch}{1.5} 
    \setlength{\tabcolsep}{6pt} 
    \resizebox{\textwidth}{!}{%
        \begin{tabular}{@{}lllS[table-format=2.2]S[table-format=1.2]S[table-format=2.2]S[table-format=4.0]@{}}
        \textbf{Dataset} & \textbf{Model} & {\makecell{\textbf{Optimizer}
        \\ \textbf{(LLM-TPE Sampler)}}} & {\makecell{\textbf{Initial Score} \\ \textbf{(CV Avg)}}} & {\makecell{\textbf{Optimized Score} \\ \textbf{(CV Avg)}}} & {\textbf{Test Score}} & {\textbf{Runtime (s)}} \\
        \midrule
        
        \multirow{4}{*}{Bike Sharing} & \multirow{2}{*}{LightGBM} 
        & Optuna & 34.7825 & 30.7747 & 60.6842 & 72 \\
        & & \textbf{Ours (Gemini-1.5-Flash)} & 30.4210 & 30.3163 & 62.8051 & 147 \\
        \cmidrule{2-7}
        & \multirow{2}{*}{XGBoost}
        & Optuna & 38.7861 & 30.5512 & 57.7546 & 69 \\
        & & \textbf{Ours (Gemini-1.5-Flash)} & 31.8870 & 29.6096 & 59.4832 & 143 \\
        
        \midrule
        
        \multirow{4}{*}{Concrete Strength} & \multirow{2}{*}{LightGBM}
        & Optuna & 10.5583 & 9.8305 & 9.7416 & 17 \\
        & & \textbf{Ours (Gemini-1.5-Flash)} & 10.7214 & 9.8419 & 9.8011 & 118 \\
        \cmidrule{2-7}
        & \multirow{2}{*}{XGBoost}
        & Optuna & 10.6025 & 10.4088 & 10.1336 & 26 \\
        & & \textbf{Ours (Gemini-1.5-Flash)} & 10.8501 & 10.0265 & 9.9726 & 88 \\
        
        \midrule
        
        \multirow{4}{*}{Energy} & \multirow{2}{*}{LightGBM}
        & Optuna & 55.3471 & 44.9908 & 113 & 73 \\
        & & \textbf{Ours (Gemini-1.5-Flash)} & 50.9820 & 43.1187 & 119 & 266 \\
        \cmidrule{2-7}
        & \multirow{2}{*}{XGBoost}
        & Optuna & 72.8360 & 42.7849 & 142.9360 & 184 \\
        & & \textbf{Ours (Gemini-1.5-Flash)} & 54.9780 & 44.7953 & 197.3804 & 277 \\
        
        \midrule

        \multirow{4}{*}{M5} & \multirow{2}{*}{LightGBM}
        & Optuna & 6.5720 & 6.2121 & 5.7726 & 315 \\
        & & \textbf{Ours (Gemini-1.5-Flash)} & 6.2194 & 6.1518 & 5.6930 & 247 \\
        \cmidrule{2-7}
        & \multirow{2}{*}{XGBoost}
        & Optuna & 7.0952 & 6.0656 & 5.9303 & 344 \\
        & & \textbf{Ours (Gemini-1.5-Flash)} & 6.2371 & 6.0500 & 5.7274 & 372 \\
        \bottomrule
        \end{tabular}%
    }
\caption{This table compares LLM-TPE Sampler with Gemini-1.5-Flash with LLM initialization and Optuna in budget-constrained settings for \textbf{regression} tasks with LightGBM and XGBoost models. For each setup, an early stopping of 5 iterations is used for regularization. LLM strategy corresponds to the reference: "llm-tpe\_sampler+gemini-1.5-flash+llm\_init+patience\_5" defined in the Methodology~\ref{methodology:llm_strategy3}.}
\label{tab:table8}
\end{table}

%% file: sections/conclusion.tex
\section{Conclusion}
This research explored the potential of large language models (LLMs) for hyperparameter optimization by comparing various models: GPT-3.5-Turbo, GPT-4o, Claude-3.5-Sonnet, and Gemini-1.5-Flash, with traditional bayesian optimization methods such as optuna and hyperopt. The study comprehensively analyzed LLMs’ performance in parameter initialization and optimization, adaptive search spaces, and the management of exploration-exploitation trade-offs. Two key contributions emerged: a novel hybrid LLM-TPE sampler for improved exploration-exploitation balance and an adaptive search space mechanism driven by LLMs to automatically evolve the parameter space. Early stopping criteria were also integrated to reduce computational costs.
Initial experiments with fully LLM-powered methods demonstrated that LLMs consistently outperformed random initialization, particularly in regression tasks, and offered competitive performance in hyperparameter optimization compared to traditional BO methods. However, challenges such as overexploitation and limited exploration arose, particularly with complex datasets.
The introduction of LangChain solved technical challenges, including memory management and request handling. It enabled longer iterations and improved performance in parameter initialization and hyperparameter tuning. While LangChain enhanced stability and reduced early stoppings, it did not entirely eliminate the problem of overexploitation.
The LLM-TPE hybrid sampler combined the LLMs' strength in initialization and exploitation with TPE’s exploration capabilities, resulting in a significantly better balance between exploration and exploitation. This method delivered superior HPO performance across multiple tasks compared to fully LLM-based and traditional BO methods. It also reduced premature early stopping and minimized computational costs through fewer LLM API calls.
Testing the LLM-TPE sampler under budget-constrained scenarios with early stopping revealed that while it remained competitive, LLM-based methods generally required more iterations to reach optimal results when subjected to strict early stopping criteria.
While this study provided a comprehensive analysis of various API-based LLMs in hyperparameter optimization, future work could extend this research by exploring the performance of open-source LLMs. This would offer broader accessibility and potentially reduce the reliance on paid APIs. Furthermore, this study focused on tabular datasets due to API-related cost constraints. Future investigations could apply the proposed methods to more complex tasks, such as image classification, segmentation, object detection, or machine translation, to evaluate their effectiveness across a broader range of domains. Lastly, it is essential to note that despite the promising results, LLMs face a significant limitation: their lack of reproducibility, as there is no guarantee of retrieving identical optimization histories across runs. 
The complete codebase for this study is available at \href{https://github.com/KananMahammadli/SLLMBO}{https://github.com/KananMahammadli/SLLMBO.}

%% file: appendices/appendix_A.tex
\section{System Messages and Prompts}
\label{appendix_A}
This appendix provides a comprehensive overview of the system messages and prompts used throughout the SLLMBO framework. These messages guide the LLM in performing hyperparameter optimization tasks, including initialization, optimization, and summarization. Each message or prompt is tailored to specific strategies and scenarios, such as Fully LLM-powered methods, LLM-TPE hybrid sampling, and reasoning-enabled optimization. 

The following sections describe these messages and prompts in detail. Each subsection begins with a brief explanation of the context and purpose, followed by the exact text of the message or prompt formatted for clarity.

\subsection{System Message for SLLMBO with Fully LLM-powered Methods and Intelligent Summary}
The system message for SLLMBO with Fully LLM-powered Methods and Intelligent Summary directs the LLM to define task objectives, search space parameters, and interaction instructions while enabling intelligent summarization of optimization history.

\begin{lstlisting}[language=, breaklines=true, tabsize=2, showtabs=false, basicstyle=\ttfamily, breakindent=0pt]
System Message:

You are an AI assistant specialized in hyperparameter tuning for machine learning models. Your goal is to optimize hyperparameters for a {model_name} model for the following problem:{problem_description} The {direction} average and minimize standard deviation of the {metric} metric used to evaluate the performance of the model.
When asked for initialization, provide: 
    1. A list of hyperparameters to tune and their 
       value ranges,if parameter is numeric, provide range as [min, max], if parameter is categorical, provide list of options.
    2. Starting values for these parameters.
        
Format your response as a JSON object with keys param_ranges' and 'initial_params'.
        
During the optimization process, you will be asked to:
    1. Decide whether to update parameter ranges or 
       continue with the last range.
    2. Suggest new parameter values based on previous 
       parameter ranges, suggested parameter values and results, and last used parameter range.
        
Balance exploration and exploitation as the optimization progresses. Keep in mind that goal is to find the best possible hyperparameter values for {model_name} model to 
{direction} the {metric} metric.
\end{lstlisting}

\subsection{System Message for SLLMBO with Fully LLM-powered Methods, Intelligent Summary, and Reasoning}
This system message extends the intelligent summary approach by requiring the LLM to provide reasoning for its parameter suggestions and search space adjustments.

\begin{lstlisting}[language=, breaklines=true, tabsize=2, showtabs=false, basicstyle=\ttfamily, breakindent=0pt]
System Message:

You are an AI assistant specialized in hyperparameter tuning for machine learning models.
Your goal is to optimize hyperparameters for a {model_name} model for the following problem {problem_description}
The {direction} average and minimize standard
deviation of the {metric} metric used to evaluate the
performance of the model.
        
When asked for initialization, provide:
    1. A list of hyperparameters to tune and their
       value ranges, if parameter is numeric, provide
       range as [min, max], if parameter is categorical, provide list of options.
    2. Starting values for these parameters.
    3. Explain reason behind your choices.
        
Format your response as a JSON object with keys 'param_ranges' and 'initial_params' and 'reason'.
\end{lstlisting}

\subsection{System Message for SLLMBO with Fully LLM-Powered Methods and LangChain}
The system message for SLLMBO with LangChain replaces manual summarization with automated memory handling. This message outlines the LLM's responsibilities in the LangChain-powered environment.

\begin{lstlisting}[language=, breaklines=true, tabsize=2, showtabs=false, basicstyle=\ttfamily, breakindent=0pt]
System Message:

You are an AI assistant specialized in hyperparameter tuning for machine learning models.
Your goal is to optimize hyperparameters for a {model_name} model for the following problem: {problem_description} 
The {direction} average and minimize standard deviation of 
the {metric} metric used to evaluate the performance of the model.
        
When asked for initialization, provide:
    1. A list of hyperparameters to tune and their 
       value ranges, if parameter is numeric, provide range as [min, max], if parameter is categorical, provide list of options.
    2. Starting values for these parameters.
        
Format your response as a JSON object with keys 'param_ranges' and 'initial_params'.
\end{lstlisting}

\subsection{System Message for SLLMBO with LLM-TPE Sampler}
This system message guides LLM on performing hyperparameter optimization with TPE sampler.

\begin{lstlisting}[language=, breaklines=true, tabsize=2, showtabs=false, basicstyle=\ttfamily, breakindent=0pt]
System Message:

You are an AI assistant specialized in hyperparameter tuning for machine learning models.
goal is to optimize hyperparameters for a {model_name} model to {direction} the {metric} metric.
Problem description: {problem_description}
        
Some trials will be suggested by Optuna's TPESampler, and their parameter suggestions and values will be shared
with you.
Always suggest parameter values within the defined ranges.
Provide your suggestions in JSON format.
Consider all previous trial results when making new suggestions.

\end{lstlisting}

\subsection{Initialization Prompt (same for all methods, except for LLM-TPE Sampler):}
Initialization prompt is used at first iteration to define hyperparameter space and suggest initial values

\begin{lstlisting}[language=, breaklines=true, tabsize=2, showtabs=false, basicstyle=\ttfamily, breakindent=0pt]
Initializaton Prompt:

Provide the initial hyperparameters to tune, their ranges, and starting values.
\end{lstlisting}

\subsection{Initialization Prompt for SLLMBO with LLM-TPE Sampler (Relative Sampler)}
The initialization prompt for the LLM-TPE Sampler guides the LLM in defining the initial parameter search space for a relative sampling strategy.

\begin{lstlisting}[language=, breaklines=true, tabsize=2, showtabs=false, basicstyle=\ttfamily, breakindent=0pt]
Initialization Prompt:

This is the first trial for hyperparameter tuning, no previous trials to consider.
Suggest based on search space: {search_space_dict}.
Suggest integer for num_leaves, max_depth, n_estimators, min_child_samples, float otherwise.
Provide only JSON format of suggested parameters with param_name: value based on parameter names from parameter ranges.
\end{lstlisting}

\subsection{Initialization Prompt for SLLMBO with LLM-TPE Sampler (Independent Sampler)}
The initialization prompt for the LLM-TPE Sampler guides the LLM in defining the initial parameter search space for a independent sampling strategy.

\begin{lstlisting}[language=, breaklines=true, tabsize=2, showtabs=false, basicstyle=\ttfamily, breakindent=0pt]
Initialization Prompt:

This is the first trial for hyperparameter tuning, no previous trials to consider.
Suggest a value for the parameter '{param_name}' within its range: {param_range}.
Provide JSON format with key and value pair: {param_name}: suggested_value
Make sure {param_name} key exists in the JSON.
\end{lstlisting}

\subsection{Optimization Prompt for SLLMBO with Fully LLM-powered Methods and Intelligent Summary}
The optimization prompt instructs the LLM to refine its suggestions and summarize optimization history intelligently.

\begin{lstlisting}[language=, breaklines=true, tabsize=2, showtabs=false, basicstyle=\ttfamily, breakindent=0pt]
Optimization Prompt:

Goal: Find the best possible hyperparameter values for
{model_name} model to {direction} the {metric} metric.
Current best score: {best_score}
Current best parameters: {best_params}
Last used parameter ranges: {current_ranges}

1. Decide whether to update the parameter ranges or      continue with the current ranges.
2. Suggest the next set of parameters to try.

Provide your response as a JSON object with keys:
    - 'update_param_ranges': boolean
    - 'new_param_ranges': dictionary of new ranges 
      (if update_param_ranges is true)
    - 'next_params': dictionary of next parameter 
       values to try

When there is no improvement to best {metric} for some iterations, consider exploration too, and balance between exploration and exploitation.
\end{lstlisting}

\subsection{Optimization Prompt for SLLMBO with Fully LLM-powered Methods, Intelligent Summary and Reasoning}
This prompt extends the optimization prompt to include detailed reasoning for parameter updates.

\begin{lstlisting}[language=, breaklines=true, tabsize=2, showtabs=false, basicstyle=\ttfamily, breakindent=0pt]
Optimization Prompt:

Goal: Find the best possible hyperparameter values for
{model_name} model to {direction} the {metric} metric.
        
Current best score: {best_score}
Current best parameters: {best_params}
Last used parameter ranges: {current_ranges}

1. Decide whether to update the parameter ranges or 
   continue with the current ranges.
2. Suggest the next set of parameters to try.
   While deciding parameter range and new parameters to try, think step by step, consider history of your choices, their results and also last parameter ranges, parameter values and scores. Consider which parameters led to the best results and which to worse results. Consider what did you learn from your mistakes and what did you learn from your successes.
        
3. Explain your choices. Whether you decided to 
   explore or exploit, and why. Whether you decided to update parameter ranges or not, and why. If you decided to update parameter ranges, why you chose the new ranges. Why did you choose the next parameters you suggested?
        
Provide your response as a JSON object with keys:
    - 'update_param_ranges': boolean
    - 'new_param_ranges': dictionary of new ranges 
      (if update_param_ranges is true)
    - 'next_params': dictionary of next parameter 
      values to try
    - 'reason': string explaining your choices

Balance between exploration and exploitation.
\end{lstlisting}

\subsection{Optimization Prompt for SLLMBO with Fully LLM-Powered Methods and LangChain}
This prompt adapts the optimization instructions to align with LangChain's memory-based approach.

\begin{lstlisting}[language=, breaklines=true, tabsize=2, showtabs=false, basicstyle=\ttfamily, breakindent=0pt]
Optimization Prompt:

Goal: Find the best possible hyperparameter values for
{model_name} model to {direction} the {metric} metric.
        
Current best score: {best_score}
Current best parameters: {best_params}
Last used parameter ranges: {current_ranges}

1. Decide whether to update the parameter ranges or 
   continue with the current ranges.
2. Suggest the next set of parameters to try.
        
Provide your response as a JSON object with keys:
    - 'update_param_ranges': boolean
    - 'new_param_ranges': dictionary of new ranges 
      (if update_param_ranges is true)
    - 'next_params': dictionary of next parameter 
       values to try

Balance between exploration and exploitation.
\end{lstlisting}

\subsection{Optimization Prompt for SLLMBO with LLM-TPE Sampler (Relative Sampler)}
The optimization prompt for the LLM-TPE sampler with relative sampling specifies the use of prior results to guide parameter updates.

\begin{lstlisting}[language=, breaklines=true, tabsize=2, showtabs=false, basicstyle=\ttfamily, breakindent=0pt]
Optimization Prompt:

Current best score: {best_score}
Current best parameters: {best_params}
Parameter ranges: {current_ranges}
Suggest based on search space: {search_space_dict}, considering all previous trials.
Suggest integer for num_leaves, max_depth, n_estimators, min_child_samples, float otherwise.
Provide only JSON format of suggested parameters with param_name: value based on parameter names from parameter ranges.
\end{lstlisting}

\subsection{Optimization Prompt for SLLMBO with LLM-TPE Sampler (Independent Sampler)}
The optimization prompt for the LLM-TPE sampler with independent sampling focuses on unbiased parameter updates.

\begin{lstlisting}[language=, breaklines=true, tabsize=2, showtabs=false, basicstyle=\ttfamily, breakindent=0pt]
Optimization Prompt:

Current best score: {best_score}
Current best parameters: {best_params}
Suggest a value for the parameter '{param_name}' within its range: {param_range}.
Provide JSON format with key and value pair: {param_name}: suggested_value
Make sure {param_name} key exists in the JSON format.
\end{lstlisting}

\subsection{Summarization Prompt for SLLMBO with Fully LLM-powered Methods and Intelligent Summary}
This prompt ensures that the LLM can summarize prior iterations to remain within token limits without losing critical information.

\begin{lstlisting}[language=, breaklines=true, tabsize=2, showtabs=false, basicstyle=\ttfamily, breakindent=0pt]
Summarization Prompt:

Summarize the conversation history, focusing on the most important information for hyperparameter tuning:
1. Keep track of the best parameters and scores found 
   so far.
2. Summarize key trends and decisions made during the 
   optimization.
3. Maintain a balance between exploration and 
   exploitation.
4. Ensure the summary is concise while retaining 
   crucial information.

Current conversation history: {conversation_history}

Provide your summary as a list of messages, each with 'role' and 'content' keys.
\end{lstlisting}

%% file: appendices/appendix_B.tex
\section{Datasets}
\label{appendix_B}
This Appendix provides detailed information about each dataset, including preprocessing, feature engineering, train-test split strategy, and features used for experimentations.

\subsection{Gas Drift Dataset}
Gas Sensor Array Drift Dataset Data Set (\href{https://archive.ics.uci.edu/dataset/224/gas+sensor+array+drift+dataset}{Dataset Link}). 20\% of the data is kept as a test set with randomly shuffled splitting, and the remaining 80\% is used during HPO for parameter evaluation.  To measure performance during HPO, ‘f1\_weighted’ is used as a metric, StratifiedKFold with five splits, randomness, and shuffling as a cross-validation splitter. Following features have been used: 'V1', 'V2', 'V3', 'V4', 'V5'. \\ Dataset summary is provided to the LLMs.
\begin{lstlisting}[language=, breaklines=true, tabsize=2, showtabs=false, basicstyle=\ttfamily, breakindent=0pt]
We are working with a tabular classification dataset for gas drift and there are 6 classes. There are only numerical features.
\end{lstlisting}

\subsection{Cover Type Dataset}
Dataset for predicting forest cover type (\href{https://archive.ics.uci.edu/dataset/31/covertype}{Dataset Link}). Only 20000 randomly chosen samples were used in this study. Test data split and cross-validation details are identical to the gas drift dataset. The following features have been used: ‘Elevation,’ ‘Aspect,’ ‘Slope,’ ‘Horizontal\_Distance\_To\_Hydrology,’ ‘And Vertical\_Distance\_To\_Hydrology.’ \\ Dataset summary is provided to the LLMs.
\begin{lstlisting}[language=, breaklines=true, tabsize=2, showtabs=false, basicstyle=\ttfamily, breakindent=0pt]
We are working with a tabular classification dataset for forest cover type prediction based on numeric features.
\end{lstlisting}

\subsection{Adult Census Dataset}
The prediction task is determining whether a person makes over 50K yearly (\href{https://archive.ics.uci.edu/dataset/2/adult}{Dataset Link}). Test data split and cross-validation details are identical to the gas drift dataset. The following features have been used: ‘age,’ ‘education-num,’ and ‘hours-per-week.’ \\ Dataset summary is provided to the LLMs.
\begin{lstlisting}[language=, breaklines=true, tabsize=2, showtabs=false, basicstyle=\ttfamily, breakindent=0pt]
We are working with a tabular classification dataset, goal is to predict binary income group of person based on age, education and working hours.
\end{lstlisting}

\subsection{Bike Sharing Dataset}
Predicting the demand for the total number of bikes for the next hour (\href{https://archive.ics.uci.edu/dataset/275/bike+sharing+dataset}{Dataset Link}). The dataset is ordered by datetime ascendingly, and the last 238 hours are kept as test data. For cross-validation-based evaluation, mean absolute error and TimeSeriesSplit with five splits and a validation size of 100 samples are used. The following features are used: ‘season,’ ‘yr,’ ‘month,’ ‘hr,’ ‘holiday,’ ‘weekday,’ ‘working day,’ ‘weathers,’ ‘temp,’ ‘temp,’ ‘hum,’ ‘wind speed.’\\ Dataset summary is provided to the LLMs.
\begin{lstlisting}[language=, breaklines=true, tabsize=2, showtabs=false, basicstyle=\ttfamily, breakindent=0pt]
We are working with a tabular regression dataset for hourly demand based on numeric datetime and weather features.
\end{lstlisting}

\subsection{Concrete Strength Dataset}
Predicting the strength of the concrete (\href{https://www.openml.org/search?type=data&status=active&id=44959}{Dataset Link}). 20\% of the dataset is sampled randomly as a test set. Mean absolute error and 5-split KFold cross-validation with shuffling are used to evaluate the performance. The following features are used: ‘cement,’ ‘blast\_furnace\_slag,’ ‘fly\_ash,’ ‘water,’ ‘superplasticizer,’ and ‘coarse\_aggregate.’\\ Dataset summary is provided to the LLMs.
\begin{lstlisting}[language=, breaklines=true, tabsize=2, showtabs=false, basicstyle=\ttfamily, breakindent=0pt]
We are working with a tabular regression dataset for concrete strength based on numeric features.
\end{lstlisting}

\subsection{Energy Dataset}
Predicting energy use of appliances in a low-energy house (\href{https://archive.ics.uci.edu/dataset/374/appliances+energy+prediction}{Dataset Link}). The dataset is ordered by datetime in ascending order, and the last 20\% of the data is kept as a test set. Cross-validation details are the same as those for the bike-sharing dataset. The following features are used: ‘lights,’ ‘T1,’ ‘RH\_1,’ ‘T2,’ ‘RH\_2,’ ‘T3,’ ‘RH\_3,’ ‘T4,’ ‘RH\_4,’ ‘T5,’ ‘RH\_5,' ‘T6,’ ‘RH\_6,’ ‘T7,’ ‘RH\_7,’  ‘T,’ ‘RH\_8,’ ‘T9,’ ‘RH\_9,’ ‘T\_out,’ ‘Press\_mm\_hg,’ ‘RH\_out,’ ‘Windspeed,’ ‘Visibility,’ ‘Tdewpoint,’ ‘rv1,’ ‘rv2’.\\ Dataset summary is provided to the LLMs.
\begin{lstlisting}[language=, breaklines=true, tabsize=2, showtabs=false, basicstyle=\ttfamily, breakindent=0pt]
We are working with a tabular regression dataset for energy consumption based on numeric features.
\end{lstlisting}

\subsection{M5 Dataset}
Predicting daily sales for multiple items (\href{https://www.kaggle.com/c/m5-forecasting-accuracy}{Dataset Link}). Considering the complexity of the dataset, necessary pre-processing and feature engineering have been used. Out of 1913 days, the dataset is considered after the 1000th day, then clustered into three groups based on the target column with KMeans Clustering, creating high, medium, and low sales item groups and 30 items selected randomly from each group. For feature engineering, day of the week, week of the year, month of the year, quarter of year, and day of year features are extracted from the daytime, and for historic sales features, last 28 to 42 days’ sales for each item, as well as the mean and standard deviation of the last 7, 14, 30, 60, 180 days’ sales shifted by 28 are added to the dataset. The reason for not using the recent 28 days is that the original M5 forecasting accuracy challenge is designed to predict each item's sales for the next 28 days. Hence, the features are added accordingly without data leakage. \\ Dataset summary is provided to the LLMs.
\begin{lstlisting}[language=, breaklines=true, tabsize=2, showtabs=false, basicstyle=\ttfamily, breakindent=0pt]
We are working with a tabular regression dataset for daily demand forecasting for multiple items at different stores.
We have categorical columns such as item, department, store ids, date related features, lag and rolling features of sales. 
In total we have 90 unique time series data, which we train together with global model.
\end{lstlisting}

%% file: main.bbl
\begin{thebibliography}{48}
\providecommand{\natexlab}[1]{#1}
\providecommand{\url}[1]{\texttt{#1}}
\expandafter\ifx\csname urlstyle\endcsname\relax
  \providecommand{\doi}[1]{doi: #1}\else
  \providecommand{\doi}{doi: \begingroup \urlstyle{rm}\Url}\fi

\bibitem[Akiba et~al.(2019)Akiba, Sano, Yanase, Ohta, and Koyama]{akiba2019optuna41}
T.~Akiba, S.~Sano, T.~Yanase, T.~Ohta, and M.~Koyama.
\newblock Optuna: A next-generation hyperparameter optimization framework.
\newblock In \emph{Proceedings of the 25th ACM SIGKDD International Conference on Knowledge Discovery \& Data Mining}, pages 2623--2631, July 2019.

\bibitem[Arden and Safitri(2022)]{arden2022comparison45}
F.~Arden and C.~Safitri.
\newblock Hyperparameter tuning algorithm comparison with machine learning algorithms.
\newblock In \emph{2022 6th International Conference on Information Technology, Information Systems and Electrical Engineering (ICITISEE)}, pages 183--188. IEEE, December 2022.

\bibitem[Bai et~al.(2023)Bai, Li, Shen, Zhang, Zhang, and Cui]{bai2023transfer8}
T.~Bai, Y.~Li, Y.~Shen, X.~Zhang, W.~Zhang, and B.~Cui.
\newblock Transfer learning for bayesian optimization: A survey.
\newblock \emph{arXiv preprint arXiv:2302.05927}, 2023.

\bibitem[Becker and Kohavi(1996)]{becker1996adult38}
B.~Becker and R.~Kohavi.
\newblock Adult [dataset], 1996.

\bibitem[Bergstra and Bengio(2012)]{bergstra2012random4}
J.~Bergstra and Y.~Bengio.
\newblock Random search for hyper-parameter optimization.
\newblock \emph{Journal of Machine Learning Research}, 13\penalty0 (2), 2012.

\bibitem[Bergstra et~al.(2011)Bergstra, Bardenet, Bengio, and K{\'e}gl]{bergstra2011algorithms3}
J.~Bergstra, R.~Bardenet, Y.~Bengio, and B.~K{\'e}gl.
\newblock Algorithms for hyper-parameter optimization.
\newblock In \emph{Advances in Neural Information Processing Systems}, volume~24, 2011.

\bibitem[Bergstra et~al.(2015)Bergstra, Komer, Eliasmith, Yamins, and Cox]{bergstra2015hyperopt42}
J.~Bergstra, B.~Komer, C.~Eliasmith, D.~Yamins, and D.~D. Cox.
\newblock Hyperopt: A python library for model selection and hyperparameter optimization.
\newblock \emph{Computational Science \& Discovery}, 8\penalty0 (1):\penalty0 014008, 2015.

\bibitem[Blackard(1998)]{blackard1998covertype37}
J.~Blackard.
\newblock Covertype [dataset], 1998.

\bibitem[Brown(2020)]{brown2020language28}
T.~B. Brown.
\newblock Language models are few-shot learners.
\newblock \emph{arXiv preprint arXiv:2005.14165}, 2020.

\bibitem[Candanedo(2017)]{candanedo2017appliances40}
L.~Candanedo.
\newblock Appliances energy prediction [dataset], 2017.

\bibitem[Du et~al.(2024)Du, Liu, Wang, Wang, Liu, Chen, and Lou]{du2024evaluating24}
X.~Du, M.~Liu, K.~Wang, H.~Wang, J.~Liu, Y.~Chen, and Y.~Lou.
\newblock Evaluating large language models in class-level code generation.
\newblock In \emph{Proceedings of the IEEE/ACM 46th International Conference on Software Engineering}, pages 1--13, April 2024.

\bibitem[Fanaee-T(2013)]{fanaee2013bike39}
H.~Fanaee-T.
\newblock Bike sharing [dataset], 2013.

\bibitem[Feurer et~al.(2014)Feurer, Springenberg, and Hutter]{feurer2014meta6}
M.~Feurer, J.~T. Springenberg, and F.~Hutter.
\newblock Using meta-learning to initialize bayesian optimization of hyperparameters.
\newblock In \emph{MetaSel@ECAI}, pages 3--10, 2014.

\bibitem[Hassanali et~al.(2024)Hassanali, Soltanaghaei, Javdani~Gandomani, and Zamani~Boroujeni]{hassanali2024software18}
M.~Hassanali, M.~Soltanaghaei, T.~Javdani~Gandomani, and F.~Zamani~Boroujeni.
\newblock Software development effort estimation using boosting algorithms and automatic tuning of hyperparameters with optuna.
\newblock \emph{Journal of Software: Evolution and Process}, page~e2, 2024.

\bibitem[Howard et~al.(2020)Howard, Makridakis, and Vangelis]{howard2020m535}
Addison Howard, Spyros Makridakis, and Vangelis.
\newblock M5 forecasting - accuracy, 2020.

\bibitem[Hutter et~al.(2019)Hutter, Kotthoff, and Vanschoren]{hutter2019automated5}
F.~Hutter, L.~Kotthoff, and J.~Vanschoren.
\newblock \emph{Automated machine learning: methods, systems, challenges}.
\newblock Springer Nature, 2019.

\bibitem[Hutter et~al.(2011)Hutter, Hoos, and Leyton-Brown]{hutter2011sequential20}
Frank Hutter, Holger~H. Hoos, and Kevin Leyton-Brown.
\newblock Sequential model-based optimization for general algorithm configuration.
\newblock In \emph{Learning and Intelligent Optimization: 5th International Conference, LION 5, Rome, Italy, January 17-21, 2011. Selected Papers}, volume~5, pages 507--523. Springer Berlin Heidelberg, 2011.

\bibitem[Joy et~al.(2020)Joy, Rana, Gupta, and Venkatesh]{joy2020fast13}
T.~T. Joy, S.~Rana, S.~Gupta, and S.~Venkatesh.
\newblock Fast hyperparameter tuning using bayesian optimization with directional derivatives.
\newblock \emph{Knowledge-Based Systems}, 205:\penalty0 106247, 2020.

\bibitem[Kasneci et~al.(2023)Kasneci, Seßler, Küchemann, Bannert, Dementieva, Fischer, and Kasneci]{kasneci2023chatgpt21}
E.~Kasneci, K.~Seßler, S.~Küchemann, M.~Bannert, D.~Dementieva, F.~Fischer, and G.~Kasneci.
\newblock Chatgpt for good? on opportunities and challenges of large language models for education.
\newblock \emph{Learning and Individual Differences}, 103:\penalty0 102274, 2023.

\bibitem[Kelly et~al.()Kelly, Longjohn, and Nottingham]{kelly2023uci33}
Markelle Kelly, Rachel Longjohn, and Kolby Nottingham.
\newblock The uci machine learning repository.
\newblock URL \url{https://archive.ics.uci.edu}.

\bibitem[Kim et~al.(2017)Kim, Kim, and Choi]{kim2017warm7}
J.~Kim, S.~Kim, and S.~Choi.
\newblock Learning to warm-start bayesian hyperparameter optimization.
\newblock \emph{arXiv preprint arXiv:1710.06219}, 2017.

\bibitem[Kojima et~al.(2022)Kojima, Gu, Reid, Matsuo, and Iwasawa]{kojima2022large29}
T.~Kojima, S.~S. Gu, M.~Reid, Y.~Matsuo, and Y.~Iwasawa.
\newblock Large language models are zero-shot reasoners.
\newblock In \emph{Advances in Neural Information Processing Systems}, volume~35, pages 22199--22213, 2022.

\bibitem[Lai et~al.(2023)Lai, Lin, Lin, Shih, Wang, and Pai]{lai2023tree43}
J.~P. Lai, Y.~L. Lin, H.~C. Lin, C.~Y. Shih, Y.~P. Wang, and P.~F. Pai.
\newblock Tree-based machine learning models with optuna in predicting impedance values for circuit analysis.
\newblock \emph{Micromachines}, 14\penalty0 (2):\penalty0 265, 2023.

\bibitem[Li et~al.(2022)Li, Shen, Jiang, Bai, Zhang, Zhang, and Cui]{li2022transfer12}
Y.~Li, Y.~Shen, H.~Jiang, T.~Bai, W.~Zhang, C.~Zhang, and B.~Cui.
\newblock Transfer learning based search space design for hyperparameter tuning.
\newblock In \emph{Proceedings of the 28th ACM SIGKDD Conference on Knowledge Discovery and Data Mining}, pages 967--977, 2022.

\bibitem[Liu et~al.(2023)Liu, Lin, Wang, Yao, Tong, Yuan, and Zhang]{liu2023large30}
F.~Liu, X.~Lin, Z.~Wang, S.~Yao, X.~Tong, M.~Yuan, and Q.~Zhang.
\newblock Large language model for multi-objective evolutionary optimization.
\newblock \emph{arXiv preprint arXiv:2310.12541}, 2023.

\bibitem[Liu et~al.(2024{\natexlab{a}})Liu, Chen, Qu, Tang, and Ong]{liu2024large31}
S.~Liu, C.~Chen, X.~Qu, K.~Tang, and Y.~S. Ong.
\newblock Large language models as evolutionary optimizers.
\newblock In \emph{2024 IEEE Congress on Evolutionary Computation (CEC)}, pages 1--8. IEEE, June 2024{\natexlab{a}}.

\bibitem[Liu et~al.(2024{\natexlab{b}})Liu, Gao, and Li]{liu2024agent16}
S.~Liu, C.~Gao, and Y.~Li.
\newblock Large language model agent for hyper-parameter optimization.
\newblock \emph{arXiv preprint arXiv:2402.01881}, 2024{\natexlab{b}}.

\bibitem[Liu et~al.(2024{\natexlab{c}})Liu, Astorga, Seedat, and van~der Schaar]{liu2024bayesian17}
T.~Liu, N.~Astorga, N.~Seedat, and M.~van~der Schaar.
\newblock Large language models to enhance bayesian optimization.
\newblock \emph{arXiv preprint arXiv:2402.03921}, 2024{\natexlab{c}}.

\bibitem[Lv(2023)]{lv2023gdbt44}
Q.~Lv.
\newblock Hyperparameter tuning of gdbt models for prediction of heart disease.
\newblock In \emph{International Conference on Electronic Information Engineering and Computer Science (EIECS 2022)}, volume 12602, pages 686--691. SPIE, April 2023.

\bibitem[Meyer et~al.(2023)Meyer, Urbanowicz, Martin, O’Connor, Li, Peng, and Moore]{meyer2023chatgpt23}
J.~G. Meyer, R.~J. Urbanowicz, P.~C. Martin, K.~O’Connor, R.~Li, P.~C. Peng, and J.~H. Moore.
\newblock Chatgpt and large language models in academia: opportunities and challenges.
\newblock \emph{BioData Mining}, 16\penalty0 (1):\penalty0 20, 2023.

\bibitem[Nguyen and Grover(2024)]{nguyen2024lico32}
T.~Nguyen and A.~Grover.
\newblock Lico: Large language models for in-context molecular optimization.
\newblock \emph{arXiv preprint arXiv:2406.18851}, 2024.

\bibitem[Nguyen et~al.(2019)Nguyen, Gupta, Rana, Li, and Venkatesh]{nguyen2019filtering10}
V.~Nguyen, S.~Gupta, S.~Rana, C.~Li, and S.~Venkatesh.
\newblock Filtering bayesian optimization approach in weakly specified search space.
\newblock \emph{Knowledge and Information Systems}, 60:\penalty0 385--413, 2019.

\bibitem[Omkari and Shinde(2022)]{omkari2022cardiovascular48}
D.~Y. Omkari and S.~B. Shinde.
\newblock Cardiovascular disease prediction using machine learning techniques with hyperopt.
\newblock In \emph{International Conference on Communication and Intelligent Systems}, pages 585--597. Springer Nature Singapore, December 2022.

\bibitem[Perrone et~al.(2019)Perrone, Shen, Seeger, Archambeau, and Jenatton]{perrone2019search11}
V.~Perrone, H.~Shen, M.~W. Seeger, C.~Archambeau, and R.~Jenatton.
\newblock Learning search spaces for bayesian optimization: Another view of hyperparameter transfer learning.
\newblock In \emph{Advances in Neural Information Processing Systems}, volume~32, 2019.

\bibitem[Poloczek et~al.(2016)Poloczek, Wang, and Frazier]{poloczek2016warm14}
M.~Poloczek, J.~Wang, and P.~I. Frazier.
\newblock Warm starting bayesian optimization.
\newblock In \emph{2016 Winter Simulation Conference (WSC)}, pages 770--781. IEEE, 2016.

\bibitem[Putatunda and Rama(2019)]{putatunda2019bayesian46}
S.~Putatunda and K.~Rama.
\newblock A modified bayesian optimization based hyper-parameter tuning approach for extreme gradient boosting.
\newblock In \emph{2019 Fifteenth International Conference on Information Processing (ICINPRO)}, pages 1--6. IEEE, December 2019.

\bibitem[Shahriari et~al.(2016)Shahriari, Bouchard-C{\^o}t{\'e}, and Freitas]{shahriari2016unbounded9}
B.~Shahriari, A.~Bouchard-C{\^o}t{\'e}, and N.~Freitas.
\newblock Unbounded bayesian optimization via regularization.
\newblock In \emph{Artificial Intelligence and Statistics}, pages 1168--1176. PMLR, 2016.

\bibitem[Snoek et~al.(2012)Snoek, Larochelle, and Adams]{snoek2012practical19}
Jasper Snoek, Hugo Larochelle, and Ryan~P. Adams.
\newblock Practical bayesian optimization of machine learning algorithms.
\newblock In \emph{Advances in Neural Information Processing Systems}, volume~25, 2012.

\bibitem[Tan et~al.(2024)Tan, Liao, Liu, Fan, Huang, Liu, and Yan]{tan2024hyperparameter1}
J.~M. Tan, H.~Liao, W.~Liu, C.~Fan, J.~Huang, Z.~Liu, and J.~Yan.
\newblock Hyperparameter optimization: Classics, acceleration, online, multi-objective, and tools.
\newblock \emph{Mathematical Biosciences and Engineering}, 21\penalty0 (6):\penalty0 6289--6335, 2024.

\bibitem[Thirunavukarasu et~al.(2023)Thirunavukarasu, Ting, Elangovan, Gutierrez, Tan, and Ting]{thirunavukarasu2023large22}
A.~J. Thirunavukarasu, D.~S.~J. Ting, K.~Elangovan, L.~Gutierrez, T.~F. Tan, and D.~S.~W. Ting.
\newblock Large language models in medicine.
\newblock \emph{Nature Medicine}, 29\penalty0 (8):\penalty0 1930--1940, 2023.

\bibitem[Vanschoren et~al.(2014)Vanschoren, Van~Rijn, Bischl, and Torgo]{vanschoren2014openml34}
J.~Vanschoren, J.~N. Van~Rijn, B.~Bischl, and L.~Torgo.
\newblock Openml: networked science in machine learning.
\newblock \emph{ACM SIGKDD Explorations Newsletter}, 15\penalty0 (2):\penalty0 49--60, 2014.

\bibitem[Vergara(2012)]{vergara2012gas36}
A.~Vergara.
\newblock Gas sensor array drift dataset [dataset], 2012.

\bibitem[Wang et~al.(2023)Wang, Yang, and Wei]{wang2023learning26}
L.~Wang, N.~Yang, and F.~Wei.
\newblock Learning to retrieve in-context examples for large language models.
\newblock \emph{arXiv preprint arXiv:2307.07164}, 2023.

\bibitem[Wang et~al.(2024)Wang, Zhu, Saxon, Steyvers, and Wang]{wang2024latent27}
X.~Wang, W.~Zhu, M.~Saxon, M.~Steyvers, and W.~Y. Wang.
\newblock Large language models are latent variable models: Explaining and finding good demonstrations for in-context learning.
\newblock In \emph{Advances in Neural Information Processing Systems}, volume~36, 2024.

\bibitem[Wang and Wang(2020)]{wang2020lightgbm47}
Y.~Wang and T.~Wang.
\newblock Application of improved lightgbm model in blood glucose prediction.
\newblock \emph{Applied Sciences}, 10\penalty0 (9):\penalty0 3227, 2020.

\bibitem[Wu et~al.(2019)Wu, Chen, Zhang, Xiong, Lei, and Deng]{wu2019hyperparameter2}
J.~Wu, X.~Y. Chen, H.~Zhang, L.~D. Xiong, H.~Lei, and S.~H. Deng.
\newblock Hyperparameter optimization for machine learning models based on bayesian optimization.
\newblock \emph{Journal of Electronic Science and Technology}, 17\penalty0 (1):\penalty0 26--40, 2019.

\bibitem[Zhang et~al.(2023)Zhang, Desai, Bae, Lorraine, and Ba]{zhang2023llm15}
M.~R. Zhang, N.~Desai, J.~Bae, J.~Lorraine, and J.~Ba.
\newblock Using large language models for hyperparameter optimization.
\newblock In \emph{NeurIPS 2023 Foundation Models for Decision Making Workshop}, 2023.

\bibitem[Zhu et~al.(2023)Zhu, Yuan, Wang, Liu, Liu, Deng, and Wen]{zhu2023large25}
Y.~Zhu, H.~Yuan, S.~Wang, J.~Liu, W.~Liu, C.~Deng, and J.~R. Wen.
\newblock Large language models for information retrieval: A survey.
\newblock \emph{arXiv preprint arXiv:2308.07107}, 2023.

\end{thebibliography}
